# REAL: REtrieval-reAsoning and Logic-constructed Attention Behaviors for Long-Context KV Cache Compression

Mengjie Li[1], Yuan Feng[2], Xike Xie[3], and William J. Song[†1]

[1]School of Electrical and Electronic Engineering, Yonsei University
[2]School of Computer Science, University of Science and Technology of China
[3]School of Biomedical Engineering, University of Science and Technology of China
lemoji@yonsei.ac.kr, yfung@mail.ustc.edu.cn, xkxie@ustc.edu.cn, wjhsong@yonsei.ac.kr

## Abstract

The growing sequence length of large language models poses significant challenges for key-value (KV) caches. Existing state-of-the-art cache eviction methods primarily analyze the inference behavior of attention heads in successful retrieval-reasoning cases, often overlooking diverse behaviors in failure cases, such as bias and distraction. This oversight limits the potential to leverage heterogeneous head behaviors for improved eviction performance. Inspired by the confusion matrix, we introduce an Attention Behavior Matrix to comprehensively analyze attention head behaviors in both success and failure scenarios. By maximizing the signal-to-noise ratio — strengthening valid reasoning pathways in success cases while inhibiting noise from bias and distraction in failure cases — we propose ***RE****trieval-re****A****soning and* ***L****ogic-constructed (REAL)* KV cache eviction, the first method to leverage multi-behavior analysis. Comprehensive evaluations show that REAL achieves remarkable performance across various models and benchmarks; notably, on LongBench v2, it achieves comparable accuracy to the strongest baseline, HeadKV-R2, while requiring 32x less space (Figure 1). By offering a novel perspective on behavior analysis, we pave the way for a shift from success-only to comprehensive, failure-aware methods in long-context modeling. Our code is available at `https://github.com/yonseicasl/REAL`.

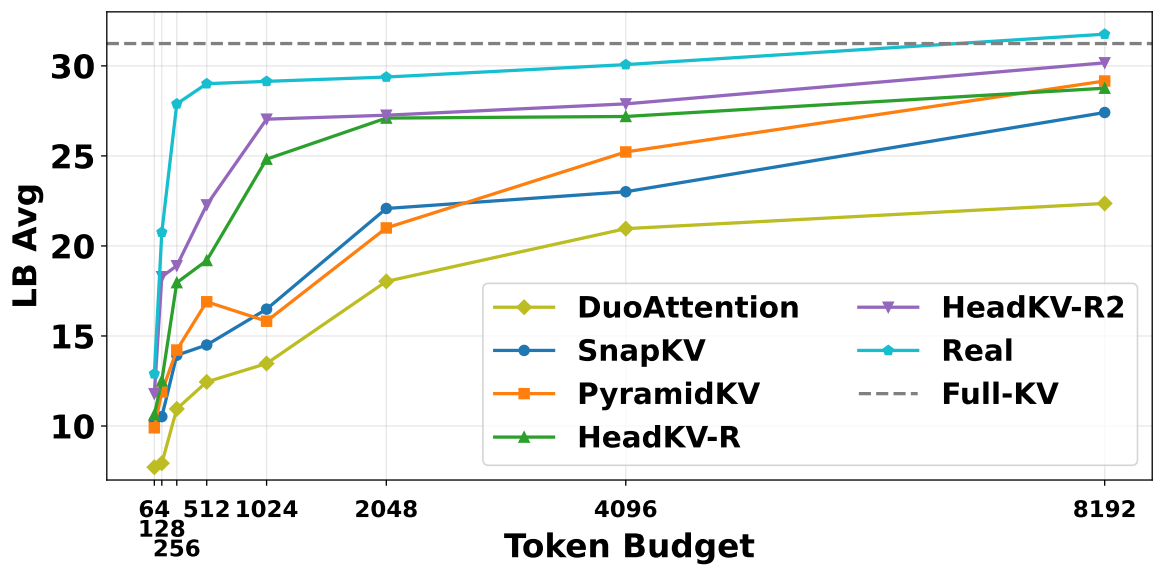

Figure 1: Results on question-aware LongBench v2 with Mistral-Large-Instruct-2411. REAL matches SOTA HeadKV-R2 using 4,096 vs. 8,192 cache tokens (2x smaller), and matches SOTA at 4,096 tokens with 128 tokens (32x smaller). See Section 4 for full results.

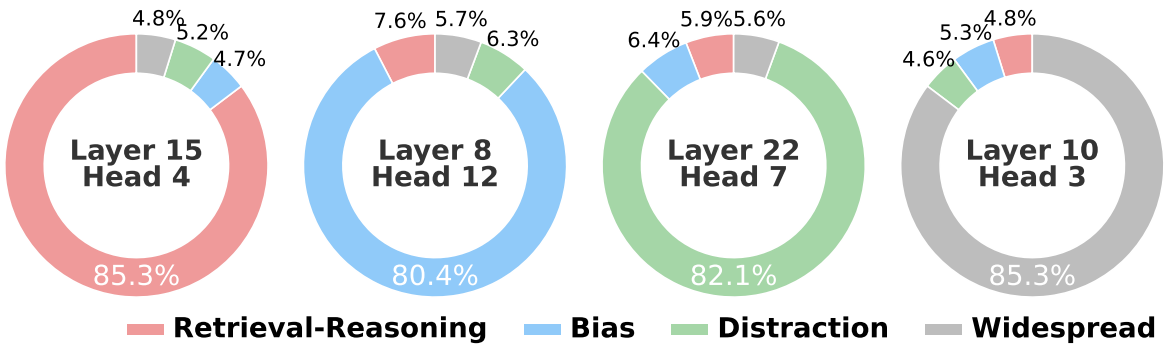

Figure 2: Heads dominated by different behaviors.

## 1 Introduction

Retrieval-driven and logic-faithful Transformer-based (Vaswani et al., 2017) large language models (LLMs) (OpenAI, 2025; Anthropic, 2025) have shown remarkable performance on tasks such as question answering (QA) (Kamalloo et al., 2023). To speed up inference, these models rely on *key-value (KV) caches*, which store key and value vectors to avoid recalculations. However, as the size of the KV cache grows with sequence length, model dimensionality, and batch size, it quickly overwhelms the memory capacity of graphics processing units (GPUs). For example, the KV cache for 64 x 4,096 tokens in GPT-3 (Brown et al., 2020) requires about 1,208 GB, whereas the device memory capacity of an NVIDIA H200 GPU is only 141 GB, underscoring the necessity of efficient KV cache compression.

To address this problem, various cache eviction methods have been proposed (Zhang et al., 2023; Cai et al., 2024; Li et al., 2024; Feng et al., 2025a,c). They enforce a fixed budget by retaining only a subset of KV entries within each attention head and discarding the rest. This reduces memory usage and speeds up decoding, enabling efficient long-context inference. However, most approaches ignore functional differences across attention heads and therefore apply uniform compression budgets, which limits the effectiveness of cache eviction. AdaKV (Feng et al., 2025b) recognizes this issue

[†]Corresponding Author.

*This author's version has been accepted for the Main Conference at the 64th Annual Meeting of the Association for Computational Linguistics (ACL 2026). The officially published version is available online under a Creative Commons Attribution 4.0 International License (CC BY 4.0). Please cite the official version.*

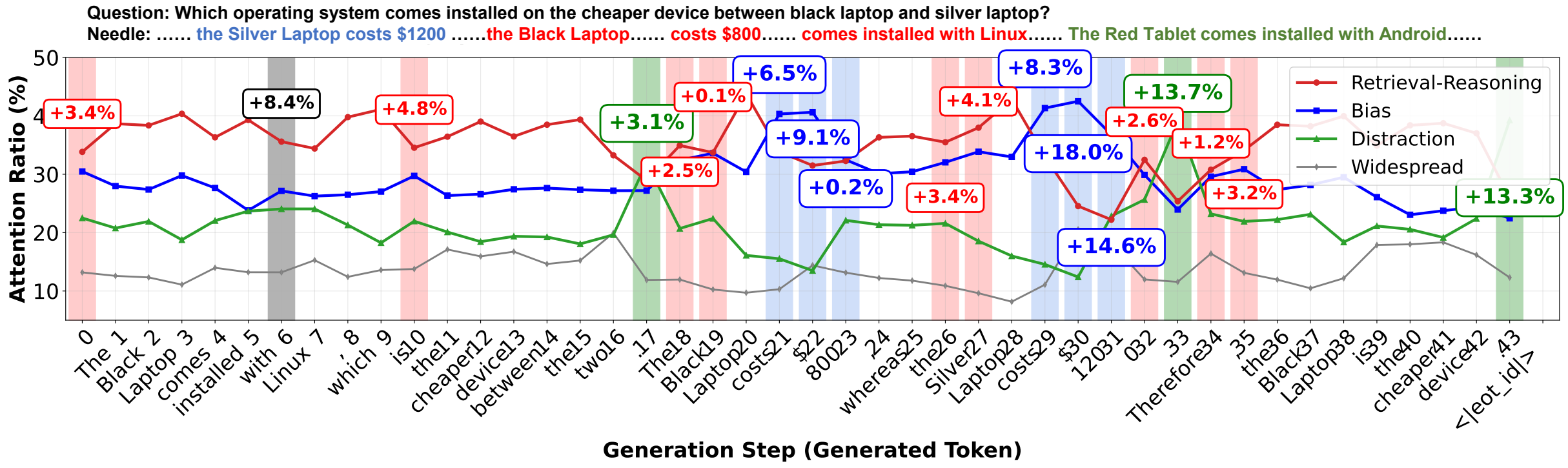

Figure 3: Dominant attention behaviors significantly affect model inference. (i) **Retrieval-Reasoning risks being misled.** At steps 0, 10, **decisive 18-19**, **decisive 26-27**, 32, 34, and 35, the lead is marginal ($< 5\%$). At the **decisive Step 6**, the advantage is merely 8.4%. (ii) **Bias causes misdirection.** At **decisive steps 21, 22, 23** for 'cheaper', the model bypassed retrieval-reasoning and extracted the price '$800' from bias. (iii) **Distraction misses insights with nearly 70%.** At two steps, 17 and 43 within three dominant steps, the model misses meaningful information.

and allocates per-head budgets based on the concentration of each head, but the gains from this coarse-grained strategy remain limited. Motivated by evidence that certain attention heads exhibit strong long-context retrieval behavior (Wu et al., 2025), subsequent methods such as DuoAttention (Tang et al., 2025; Xiao et al., 2025) probe these retrieval heads using synthetic data and assign them larger budgets during cache eviction. HeadKV (Fu et al., 2025) further extends this work by analyzing both retrieval and reasoning behaviors.

However, these methods focus exclusively on successful cases, where the model's prediction matches the ground truth, and therefore overlook the diverse head behaviors that arise when inference fails. Inspired by confusion-matrix analysis, we show that head behavior can be categorized into four cases in Table 1; beyond the previously studied success cases, three additional categories remain during LLM inference (Figure 3). Specifically, biased attention can mislead the model, yielding a false positive (FP), while distracted attention can miss relevant evidence, resulting in a false negative (FN). We therefore argue that efficient cache eviction requires a dual strategy: maximizing the signal-to-noise ratio (Shannon, 1948) by strengthening valid reasoning pathways while inhibiting the noise arising from bias and distraction. As illustrated in Figure 2, different heads exhibit distinct behaviors within these failure cases (i.e., bias and distraction), revealing significant potential for optimization through fine-grained, head-wise budget allocation.

In this paper, we propose ***RE**trieval-re**A**soning and **L**ogic-constructed (REAL)*, the first method to perform fine-grained behavior-aware budget allocation across heads. Drawing on confusion-matrix–derived metrics such as precision, recall, and F1-score, we introduce three metrics: *Retrieval-reAsoning score (RAsc)*, *Logic-Constructed score (LCsc)*, and *INFerence score (INFsc)* based on the above attention behavior matrix. High *INFsc* is achieved when both RAsc and LCsc are simultaneously high, which means the head attends more to retrieval-reasoning and less to bias and distraction. Such heads are allocated a larger KV budget (Section 3.1). Extensive experiments show that REAL achieves state-of-the-art performance across question-aware, question-agnostic, and non-QA scenarios. For example, as shown in Figure 1, on the challenging LongBench v2 benchmark, REAL matches HeadKV-R2 using a cache budget of 4,096 instead of 8,192 (2x fewer), and achieves comparable performance with as little as 128 cache budget (32x fewer than 4,096). The following summarizes key contributions:

| | | **Ground Truth** | |
|---|---|---|---|
| | | **Answer Related (P)** | **Non-Answer Related (N)** |
| **Inference** | **Answer Related (P)** | Retrieval-Reasoning (TP) | Bias (FP) |
| | **Non-Answer Related (N)** | Distraction (FN) | Widespread (TN) |

Table 1: Attention behavior matrix: A complete diagnostic framework inspired by the confusion matrix.

- We identify a key limitation of existing methods: the neglect of attention behaviors during inference failure, which severely bottlenecks head-wise cache eviction performance.
- We propose a complete Attention Behavior Matrix inspired by confusion matrices, introducing three derived metrics — *RAsc*, *LCsc*, and *INFsc* — to quantify head-wise attention patterns.

- We present *REAL*, the first comprehensive attention-behavior–aware framework, delivering consistent and substantial improvements across comprehensive experiments.
- Our work establishes a new multi-behavior head analysis perspective, encouraging a shift beyond single-pattern, success-only head labeling toward comprehensive, failure-aware methods for long-context inference.

## 2 Preliminary

In this section, we describe (i) Manual needles' rationality, (ii) How to label a needle, (iii) Construction of the attention behavior matrix, (iv) Computation of the *RAsc*, *LCsc*, and *INFsc* metrics, and (v) Sensitivity analysis of *INFsc* and KV allocation.

### 2.1 Manually Constructed Needles' Rationality

Manual probing has emerged as a well-established methodological framework, as demonstrated by recent studies (Wu et al., 2025; Fu et al., 2025). These works underscore the robustness and efficacy of this paradigm in long-context reasoning. Appendix A.6 further analyzes retrieval-reasoning difficulty and correlates synthetic and task-extracted needles, confirming the utility of hand-crafted needles.

### 2.2 Needle Labeling

The labeling procedure is robust across multiple needle generators and alternative constructions. WR/WB/WD labels remain perfectly stable with respect to generator-specific phrasing, as they depend solely on sentence-level logical roles. The explicit construction is detailed in Table 2.

| | Question-Related | Answer-Related | Similar to Answer Source Structure |
|---|---|---|---|
| **Retrieval & Reasoning** | ✓ | ✓ | N/A |
| **Bias** | ✓ | ✗ | ✓ |
| **Distraction** | ✗ | ✗ | ✓ |
| **Widespread** | ✗ | ✗ | ✗ |

Table 2: Properties of information labels.

### 2.3 Attention Behavior Matrix

Referencing the experimental setup (Wu et al., 2025), we manually construct novel needle cases that the model has never encountered, as shown in Table 3. Each distinct attention behavior serves a unique role in long-context processing. Retrieval-Reasoning captures the core question-answer relationship by accurately identifying and attending to answer-relevant tokens. It exhibits high true positive rates, demonstrating its ability to retrieve critical information that directly addresses the query. Bias focuses on question-related context rather than answer sources, which often leads to false positive references. Distraction exhibits structural sentence similarity to Retrieval-Reasoning behavior but fails to capture the actual answer content. It yields high false negative rates by missing genuinely relevant information. Widespread behavior maintains diffuse attention across the entire context, simulating pervasive background information processing. It shows high true negative rates, indicating its role in broadly monitoring the context without selectively focusing on specific tokens.

The design ensures that the model relies on the KV cache to get knowledge for inference, rather than falling back on internally stored knowledge learned during pre-training. We sampled 30 different sequence lengths, ranging from 1K to 30K tokens, in steps of 1,024. For each sequence length, the query was inserted at 33 uniform positions between 2% and 98%, in steps of 3%.

### 2.4 Inference Score Calculation

We construct the attention confusion matrix in Table 1 based on the *dominant attention behavior*, which is determined by the argmax of the four cumulative needle attention weights per head at each generation step. Analogous to precision, recall, and F1-score (Susmaga, 2004), we define *Retrieval-reAsoning score (RAsc)*, *Logic-Constructed score (LCsc)*, and *INFerence score (INFsc)* as Equations (1), (2), and (3), where $W_{\mathrm{R}}$, $W_{\mathrm{B}}$, and $W_{\mathrm{D}}$ denote attention weights to retrieval-reasoning, bias, and distraction, respectively. High *INFsc* is achieved when both *RAsc* and *LCsc* are high, which means $W_{\mathrm{R}}$ is high, $W_{\mathrm{D}}$ and $W_{\mathrm{B}}$ are low. In this state, the head focuses more on retrieval-reasoning and pays less attention to bias and distraction, and should therefore be allocated a larger KV budget.

$$RAsc = \frac{W_{\mathrm{R}}}{W_{\mathrm{R}} + W_{\mathrm{D}}}, \tag{1}$$

$$LCsc = \frac{W_{\mathrm{R}}}{W_{\mathrm{R}} + W_{\mathrm{B}}}, \tag{2}$$

$$INFsc = \frac{2 \cdot RAsc \cdot LCsc}{RAsc + LCsc} \tag{3}$$

### 2.5 Sensitivity for *INFsc* and Allocation

**Surface semantics.** *INFsc* and thus KV allocation are largely invariant to specific phrasing, templates,

| Method | Example |
|---|---|
| **Retrieval (HeadKV-R)** | **Question:** What is the best thing to do in Beijing?<br>**Needle:** The best thing to do in Beijing is to take a walk in Chaoyang Park and have a cup of Espresso in the evening. (k part) |
| **Retrieval-Reasoning (HeadKV-R2)** | **Question:** What is the favorite thing of the younger one between John and Mary?<br>**Needle:** John is 12 years old. Mary is 13 years old. (r part) Mary's favorite thing is to take a walk in Chaoyang Park and have a cup of Espresso in the evening. ($c^1$ part) John's favorite thing is to play basketball at the local gym and enjoy a smoothie afterward. ($c^2$ part) |
| **Attention Behavior (REAL)** | **1. Question:** Which operating system comes installed on the cheaper device between the black laptop and the silver laptop?<br>**Needle:** Reflecting the 2025 industry shift towards AI-integrated computing and neural processing, the Silver Laptop costs $1200, whereas the Black Laptop, targeting the essential productivity market without NPU acceleration, costs $800. The Black Laptop comes installed with Linux. The Silver Laptop comes installed with Windows. Microsoft Windows is known for its graphical user interfaces and broad hardware compatibility. The Red Tablet comes installed with Android. Google Android is based on the Linux kernel and designed primarily for mobile devices.<br>**2. Question:** Where is the earlier meeting held between Budget Review and Team Building?<br>**Needle:** The Budget Review, which will focus on the Q3 fiscal analysis and cost-cutting strategies, is scheduled for 9:00 AM, while the Team Building event starts at 2:00 PM. The Budget Review is held in Conference Room A. The Team Building is held in Conference Room B on the second floor, aiming at fostering cross-departmental collaboration and morale and fostering collaboration, communication, and trust among colleagues. The Yoga Class is held in Conference Room C on the third floor at 9:00 PM, providing an opportunity for employees to relax and rejuvenate for physical and mental well-being.<br>**Attention Behavior:** Retrieval-Reasoning, Bias, Distraction, and Widespread |

Table 3: Comparison of needle example. In HeadKV-R (Fu et al., 2025), the correct answer is retrieved from $k$. In HeadKV-R2 (Fu et al., 2025), the correct answer is derived from $c^2$ given background $r$, while the influence of misleading $c^1$ is **neglected**. In REAL, the correct answer is obtained by strengthening retrieval-reasoning behavior, while inhibiting the influence of bias and distraction. More cases can be seen in Appendix A.6.

and topical content. Manual needles explicitly identify the model's intrinsic logical reasoning capabilities. For instance, as illustrated in Example 2 of Table 3, identifying the keyword "earlier" requires a temporal magnitude (9:00 AM, 2:00 PM, 9:00 PM), a logical reasoning process that remains consistent despite syntactic variations, including passive voice ("is scheduled"), active voice ("starts"), or prepositions ("at"). Consequently, *INFsc* consistently prioritizes the same functional attention heads responsible for numerical and temporal comparison, ensuring that the allocation strategy remains robust to phrasing perturbations.

**Heuristics.** *INFsc* and thus KV allocation are almost insensitive to the heuristics used to classify tokens or sentences, because the essence of being "answer-related" is fundamentally based on a strict logical relationship ($A \to B \to C$), rather than fragile, fragmented token-level jitter. As Figure 3 demonstrates, the reasoning excludes explicit token matching (e.g., token "cheaper" never appears in the needle). The reasoning path follows a three-step process: (i) *Comparing* numerical values ($800 vs. $1200) $\to$ cheaper, (ii) *Entity mapping* cheaper $\to$ Black Laptop, and (iii) *Semantic synthesizing* cheaper Black Laptop + operating system $\to$ Linux. Consequently, *INFsc* consistently prioritizes the same attention heads responsible for this underlying logical inference, ensuring the resulting allocation is immune to any surface-level heuristics.

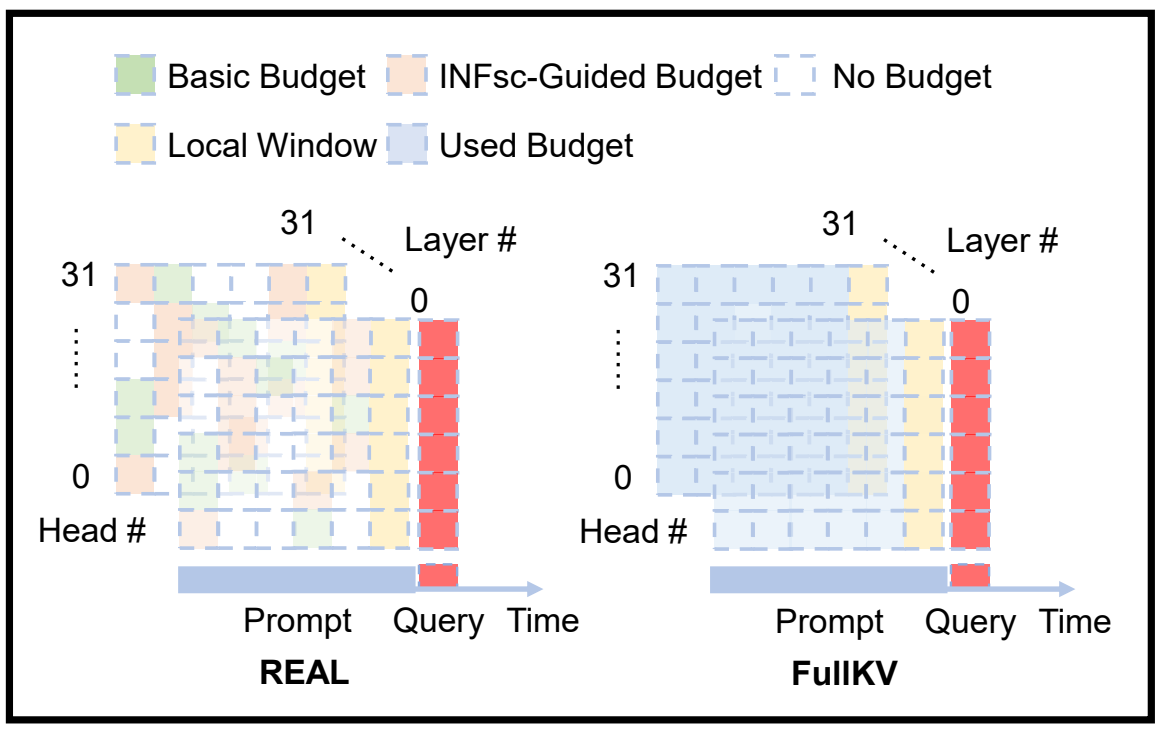

Figure 4: KV budget allocation.

## 3 KV Budget Allocation Method

In this section, we describe (i) How to allocate the KV budget referring to INFsc and (ii) How to select the KV entry within the KV budget.

### 3.1 KV Budget Allocation Across Heads

The illustration of our head-wise KV cache allocation is shown in Figure 4. Algorithm 1 compresses the KV cache via top-$k$ selection and flattened cache updates. The fixed token budget $b$ can be replaced by a ratio $\ell r$. The procedure first

**Algorithm 1** Per-head KV budget allocation.

**Input**: initially assigned cache ratio $r$ for a head, sequence length $\ell$, total budget $b_c$, predefined allocation ratio $\beta$, head $(i, j)$, layer count $L$, head count per layer $H$, INF score $H_{i,j}^{\text{inf}}$

**Output**: capacity $\mathrm{C}_{i,j}$

1: Basic budget $b_{\text{base}} \leftarrow \ell r \left(1 - \frac{1}{\beta}\right)$

2: Global budget pool $B_{\text{total}} \leftarrow \frac{\ell r}{\beta} L\, H$

3: Dynamic allocation $b_{i,j}^{\text{dyn}} \leftarrow B_{\text{total}} \cdot H_{i,j}^{\text{inf}}$

4: $b_{i,j} \leftarrow b_{\text{base}} + b_{i,j}^{\text{dyn}}$

5: **return** $\mathrm{C}_{i,j} \leftarrow \lfloor \max(0,\, b_{i,j}) + 0.5 \rfloor$

checks whether the KV length exceeds the budget. If it does not, the original $K$ and $V$ are returned unchanged. If it does, the full query window is retained, and the top-$k$ most relevant historical tokens within the budget are selected to form a compressed KV. Notably, the variable $r$ is the optimization target and not counted as a hyperparameter.

### 3.2 KV Selection with Allocated Budget

For a given head, if the sequence length remains within the budget, all KV states are retained. When the KV length exceeds its allocated budget, the KV pairs from the most recent query window are first preserved to maintain generation coherence in Equations (4) and (5). $S$ is the attention score computed from the query window $Q_w$ and cached keys $K_c$ according to (Li et al., 2024; Cai et al., 2024; Feng et al., 2025b). $KV_{\text{res}}$ represents the compressed KV cache, comprised of the fully retained KV states from the latest query window $KV_w$ and the top-$k$ entries selected from the previous cache. The parameter $r$ denotes the cache ratio. Figure 5 illustrates the resulting KV budget distribution.

$$S(Q_w, K_c) = \text{Softmax}\left(\frac{Q_w K_c^T}{\sqrt{d_k}}\right) \quad (4)$$

$$KV_{\text{res}} = \text{Gather}(K_c, \text{TopK}(S, r)) + KV_w \quad (5)$$

## 4 Experiments and Analysis

We conduct comprehensive experiments to evaluate REAL's effectiveness in KV cache compression across multiple dimensions. (i) Experimental setup: We detail the evaluation framework, including model architectures, benchmark datasets, baseline methods, and two compression scenarios with different budget constraints. (ii) Main results: We present performance comparisons under question-aware compression (evaluated with fixed token budgets) and question-agnostic compression (evaluated with cache ratios), demonstrating REAL's superior efficiency across various compression levels. (iii) Catastrophic analysis via inverted allocation. (iv) Ablation studies: We systematically examine the contribution of each attention head behavior component and validate the effectiveness of our KV allocation guidance metric *INFsc*. (v) Hyperparameter analysis: We analyze the impact of $\beta$, the only hyperparameter, demonstrating REAL's performance across different settings. We provide detailed results and visualizations to illustrate how behavior-aware head selection improves long-context understanding.

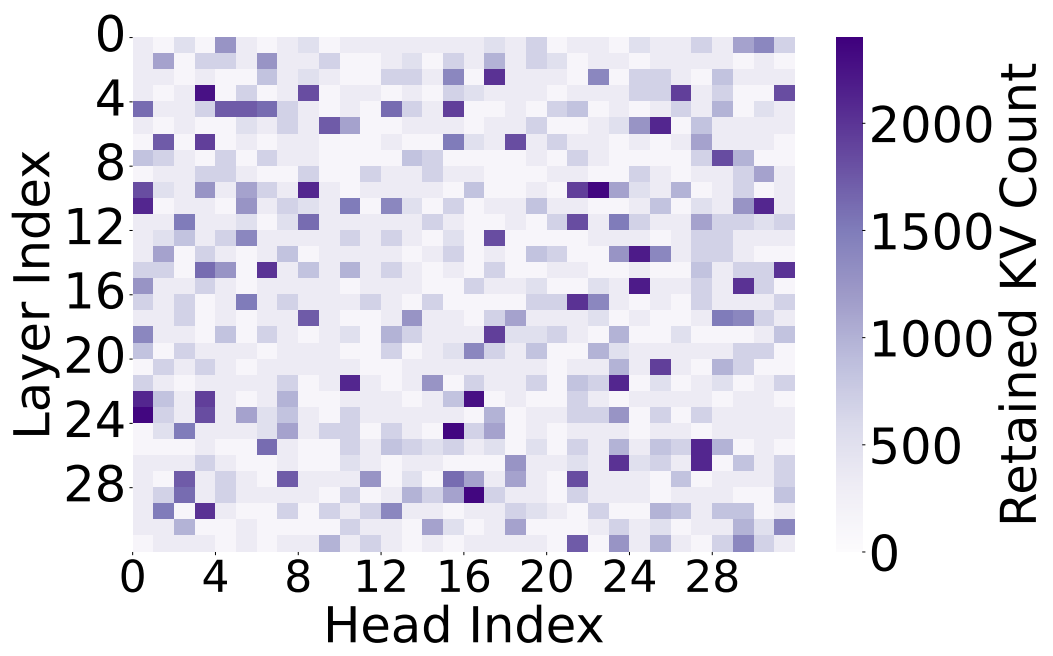

Figure 5: Retained KV counts when token budget = 512 for Llama-3-8B-Instruct (Grattafiori et al., 2024) on the 18K-length NarrativeQA dataset.

### 4.1 Settings

**Base models.** We conduct experiments on three models with maximum context lengths ranging from 8K to 128K: Llama-3-8B-Instruct (Grattafiori et al., 2024), Mistral-7B-Instruct (Jiang et al., 2023), and 123B Mistral-Large-Instruct-2411 (Mistral AI, 2024).

**Datasets.** Evaluations are performed using two benchmarks. LongBench (Bai et al., 2024) covers 16 long-context, knowledge-intensive subsets across six tasks: multi-document QA, single-document QA, summarization, few-shot learning, synthetic reasoning, and code, with sequence lengths ranging from 1K to 18K tokens. LongBench v2 (Bai et al., 2025) extends this to 20 subsets across six tasks, covering long in-text learning, long-dialogue history understanding, and long structured data understanding, with sequence lengths from 14K to 167K tokens. A detailed description is in Appendix A.4. For evaluation, we select 27 datasets from these two benchmarks.

**Baselines.** We consider existing methods with different granularity: SnapKV (Li et al., 2024), PyramidKV (Cai et al., 2024), DuoAttention (Xiao et al., 2025), HeadKV-R (Fu et al., 2025), and HeadKV-R2 (Fu et al., 2025), all under the same token bud-get and cache ratio for fair comparison.

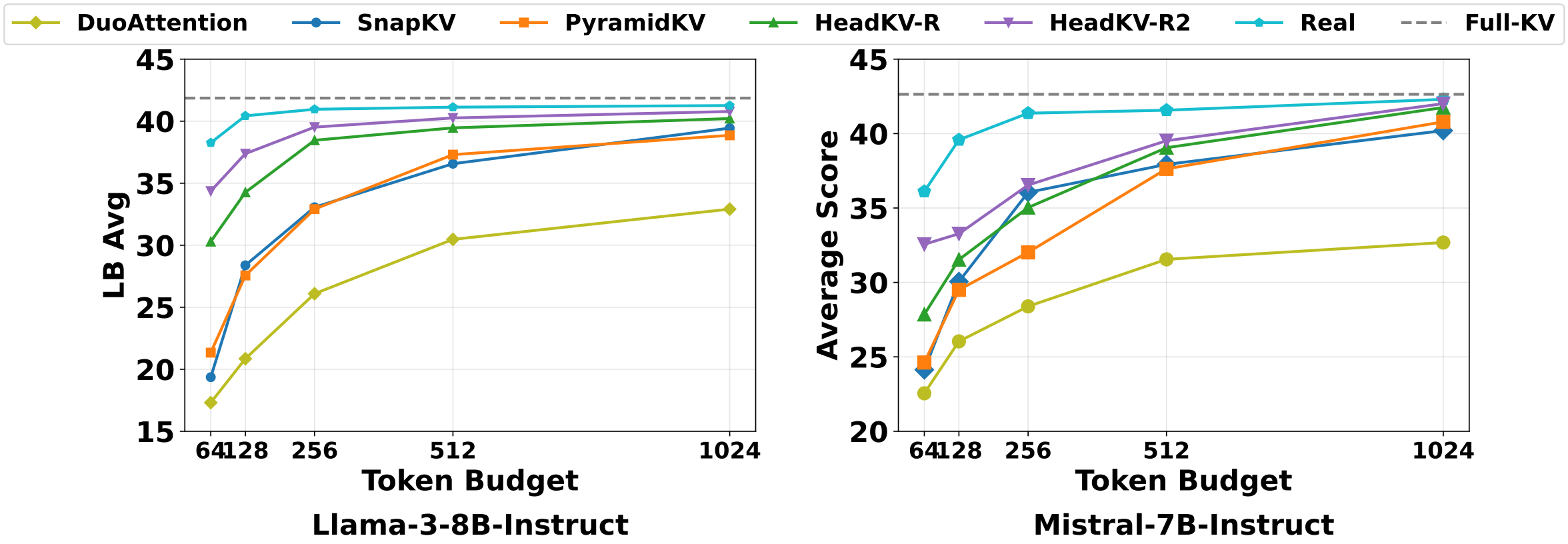

Figure 6: Question-aware performance comparison on LongBench (Bai et al., 2024). REAL outperforms all baseline methods across token budgets (64-1024) for both Llama-3-8B-Instruct (Grattafiori et al., 2024) and Mistral-7B-Instruct (Jiang et al., 2023). Comprehensive results are provided in Appendix A.7.

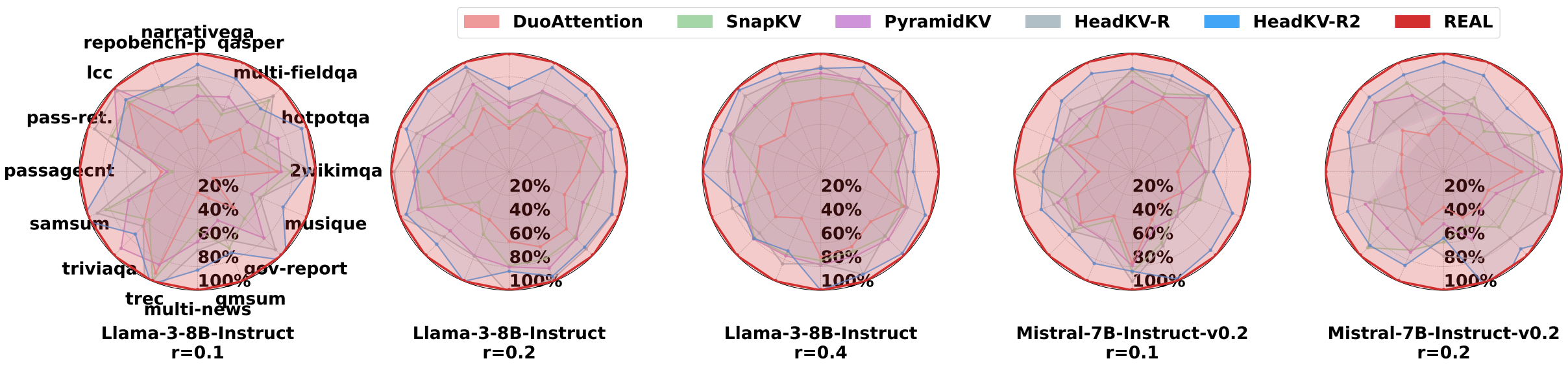

Figure 7: Question-agnostic radar chart comparison of KV cache compression methods across 16 LongBench (Bai et al., 2024) tasks. Performance is normalized to REAL (100%) for Llama-3-8B-Instruct (Grattafiori et al., 2024) and Mistral-7B-Instruct (Jiang et al., 2023) at different cache ratios (r=0.1, 0.2, 0.4). REAL consistently outperforms all baseline methods across tasks and compression levels. Comprehensive results are provided in Appendix A.7.

**Compression scenarios.** We evaluate two compression scenarios: (i) question-aware compression (Li et al., 2024), where questions are compressed alongside prefix context, enabling targeted KV cache reduction for specific queries and (ii) question-agnostic compression (Feng et al., 2025b), where only the prefix context is compressed without knowing future questions, representing more realistic and challenging cases. The former is evaluated under fixed token budgets (e.g., 128, 4096, 8192 tokens), while the latter uses cache ratios (e.g., retaining 10%, 20% of the sequence length).

**Technical details.** By monkey-patching the forward passes, REAL maintains a flattened KV cache that is initialized during prefilling and appended during decoding. To avoid `torch.cat` overhead, a specialized `update_flatten_view` CUDA kernel directly appends decoding tokens to the flattened KV cache. This streamlined memory management approach allows for a rigorous assessment of compression efficiency, demonstrating that REAL substantially reduces memory footprints without compromising inference speed.

### 4.2 Main Result

**Question-aware compression.** Under the fixed token budget constraint, Figures 1 and 6 demonstrate the superior performance of REAL across 16 datasets, underscoring the effectiveness of the proposed comprehensive attention behavior analysis. On Llama-3-8B-Instruct (Grattafiori et al., 2024), REAL achieves approximately 98% of Full-KV accuracy with a budget of 256 tokens. On Mistral-7B-Instruct (Jiang et al., 2023), REAL maintains its advantage, achieving approximately 98.6% of Full-KV's accuracy at budget 512. Most notably, on LongBench v2 (Mistral AI, 2024), REAL not only approaches Full-KV performance but surpasses it by 0.6 points at budget 8192 (31.8 vs. 31.24), demonstrating that REAL can enhance model performance through improved KV budget allocation. At budget 2048, REAL achieves 93% of Full-KV.

**Question-agnostic compression.** We present comprehensive radar chart comparisons across 16 LongBench tasks in Figure 7, showing performance on Llama-3-8B-Instruct (Grattafiori et al., 2024) ($r$=0.1, 0.2, 0.4) and Mistral-7B-Instruct-v0.2 (Jiang et al., 2023) ($r$=0.1, 0.2), normalized relative to REAL (100%). REAL consistently achieves the best performance across all tasks and cache ratios, forming the outermost boundary. Under extreme compression ($r$=0.1), REAL significantly outperforms all baselines, while DuoAttention and SnapKV reach only 40-60% of REAL's performance on most tasks. HeadKV-R2 demonstrates the second-best performance at 70-90% of REAL. As the cache ratio increases, the performance gap narrows, but REAL maintains its advantage, particularly on challenging tasks such as TriviaQA, TREC, and LCC. These results indicate that REAL's attention behavior-aware approach provides consistent improvements across diverse compression scenarios. Consequently, success-failure awareness enables more resilient, question-agnostic compression by hedging against information loss in the absence of explicit query guidance.

| Logic | Inverted Prediction | Correct Answer |
|---|---|---|
| Entity | Poetry. | The Atlas Mountains. |
| Why? | Life is but a walking shadow. | Because she is in unrequited love with someone else. |

Table 4: Representative instances showing the model hallucinates irrelevant text instead of logical answers.

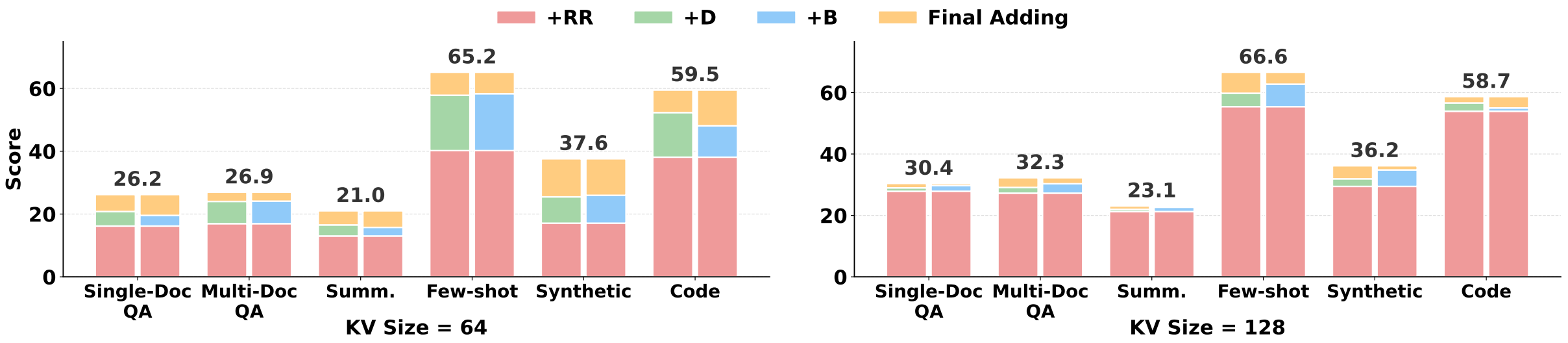

Figure 8: Component-wise ablation. The adding performance validates the complementary information for effective KV cache allocation by three components. Comprehensive results are provided in Appendix A.7.

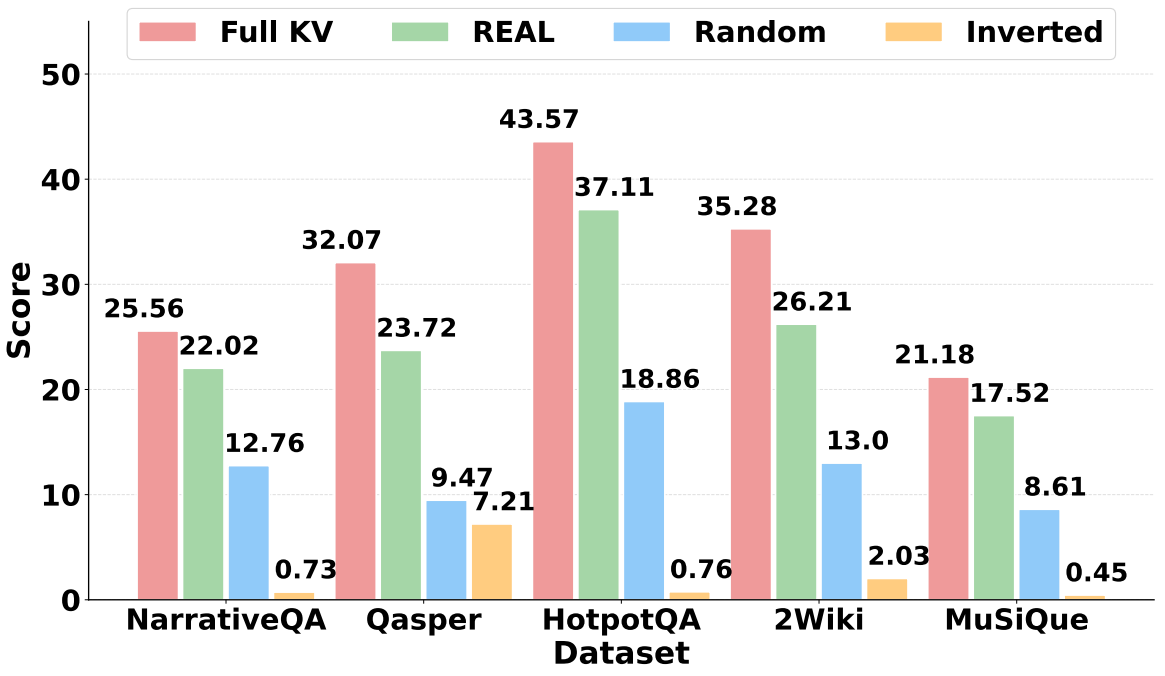

Figure 9: Measurable results showing severe accuracy degradation from inverted allocation under 64 token budget with Llama-3-8B-Instruct (Grattafiori et al., 2024).

## 4.3 Catastrophe via Inverted Allocation

To further validate the correlation between *INFsc* and model performance, we evaluate an inverted KV allocation strategy on Llama-3-8B-Instruct (Grattafiori et al., 2024) with a fixed budget of 64. In contrast to the REAL strategy, this approach intentionally prioritizes heads with the lowest *INFsc*, which are inherently more susceptible to bias and distraction within the question-aware context. As shown in Table 4, the inverted prioritization leads to a significant performance collapse. This qualitative degradation is further elucidated by the quantitative results in Figure 9.

These results confirm the critical importance of high-*INFsc* heads for the retrieval-reasoning of task-relevant information and the maintenance of logical consistency.

## 4.4 Ablation Study

**Attention behavior ablation.** We investigate the contribution of each attention behavior component in Figure 8. The performance gains are most pronounced on Few-shot tasks (from +RR 40.2 to 65.2 for KV64, and from 55.4 to 66.6 for KV128) and Code tasks (from 38.1 to 59.5 for KV64, and from 53.9 to 58.7 for KV128), whereas Synthetic tasks exhibit minimal improvements. Both bias and distraction considerations contribute meaningfully, with bias providing slightly larger gains.

**Metric ablation.** We conduct ablation studies on different score aggregation strategies, as shown in Figure 11, comparing Random_score, Mean_score, Max_score, and *INFsc* (REAL) across six task categories under token budgets of 64 and 128 on LongBench (Bai et al., 2024). REAL consistently achieves the highest retention rates, reaching 90% on Few-shot tasks for budget 64 and 99% for budget 128, significantly outperforming Random_score, which attains only 46% and 53% respectively. Max_score shows intermediate performance at 66% for budget 64 and 81% for bud-get 128 on Few-shot tasks, while Mean_score achieves 58% and 70% respectively. The performance gap is most pronounced on Few-shot and Code tasks, demonstrating that INFsc's attention behavior-aware scoring significantly outperforms simple aggregation methods, especially under tighter budget constraints.

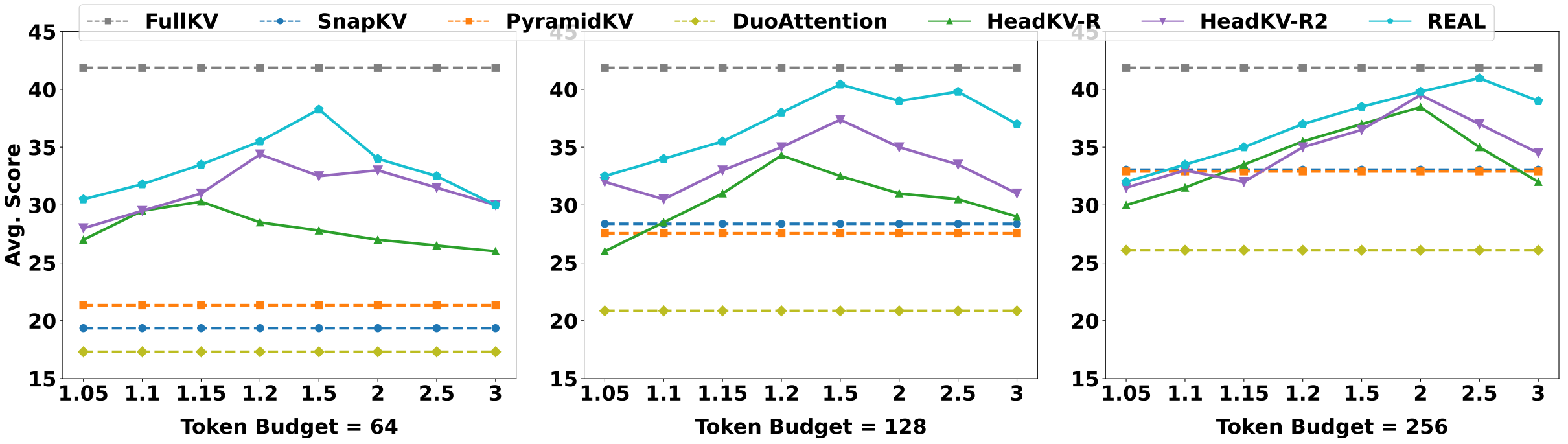

Figure 10: Hyperparameter sensitivity of the allocation parameter $\beta$ across different token budgets on Llama-3-8B-Instruct (Grattafiori et al., 2024). REAL demonstrates robust performance with optimal $\beta$ around 1.5-2.5, while HeadKV-R and HeadKV-R2 (Fu et al., 2025) show higher sensitivity. Even with sub-optimal $\beta$ settings, REAL consistently maintains a substantial performance margin. Comprehensive results are in Appendix A.7.

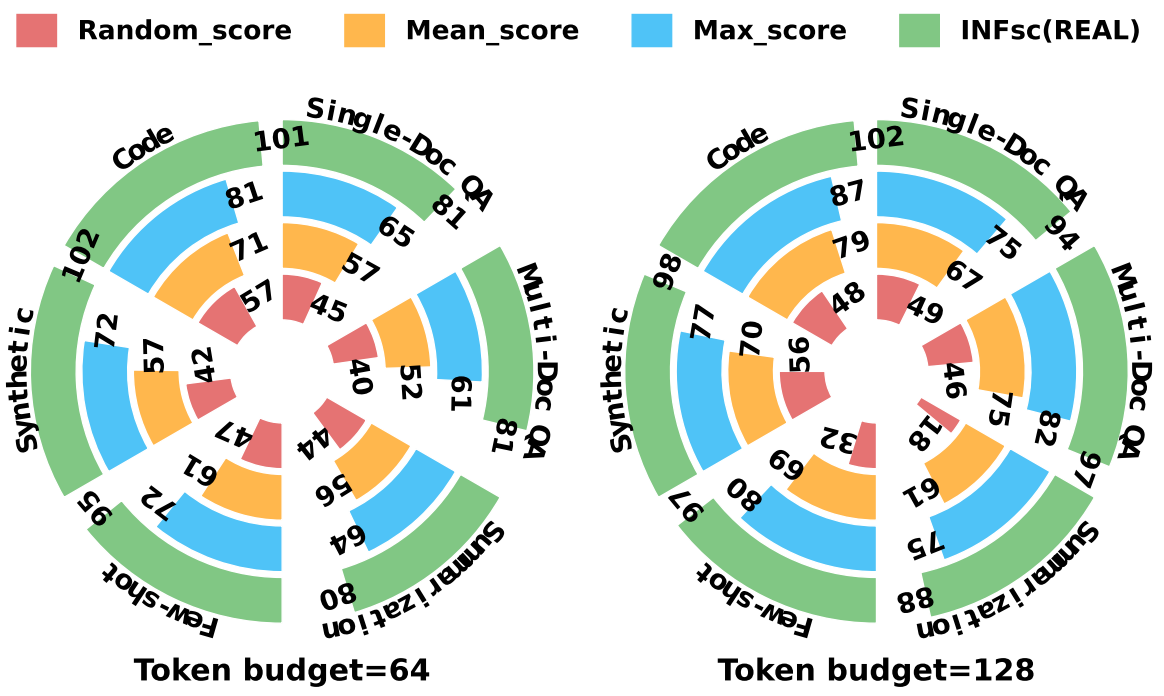

Figure 11: Performance retention of different metrics. REAL (*INFsc*) achieves superior performance compared to Random_score, Mean_score, and Max_score under token budgets of 64 (left) and 128 (right), with particularly strong performance on synthetic and code tasks. Comprehensive results are provided in Appendix A.7.

Metrics such as Random_score, Mean_score, and Max_score often rely on conventional statistical importance. In contrast, by evaluating heads based on their necessity for a successful reasoning path, REAL guarantees sufficient caching for failure-preventing heads while minimizing the footprint of those that are peripheral to the task logic.

### 4.5 Hyperparameter Analysis

REAL introduces only one hyperparameter, $\beta$, which controls the size of the global shared budget pool $B$. We select $\beta \in \{1.05, 1.1, 1.15, 1.2, 1.5, 2, 2.5, 3\}$, as shown in Figure 10. A well-calibrated $\beta$ enables dynamic resource distribution by sharing the global KV budget. This flexibility effectively captures the most salient attention heads across varying sequence lengths and diverse model architectures on different tasks.

## 5 Resource Analysis

To comprehensively evaluate the practical efficiency of REAL, we examine three key resource metrics: (i) Time-to-first-token (TTFT) latency, a critical factor for user utilization that measures the delay before generation, (ii) The computational overhead of the attention behavior matrix, which quantifies the one-time cost of our behavior-aware KV allocation for a single head, and (iii) Peak memory consumption during inference, which directly affects deployment feasibility on resource-constrained devices. Our results show that REAL achieves substantial memory reduction with negligible latency overhead.

**Time to first token (TTFT).** The latency is measured using the Reasoning-in-a-Haystack dataset. After the model finishes encoding the input sequence, REAL performs KV cache compression. We conduct 10 iterations and calculate the average running latency for the first token. The results are summarized in Table 5. Compared with Full-KV, REAL introduces only a minimal extra time cost and does not noticeably increase TTFT. In contrast, the other baselines incur substantially higher latency overhead.

**Attention behavior matrix overhead.** To evaluate the practical feasibility of our approach, we measure the computational overhead of constructing attention behavior matrices. Table 6 shows that the computational overhead is negligible, with average runtimes of only 26.04ms for Llama-3-8B-Instruct (Grattafiori et al., 2024) and 27.86ms for Mistral-7B-Instruct-v0.2 (Jiang et al., 2023), significantly

| Method | Iter.1 | Iter.2 | Iter.3 | Iter.4 | Iter.5 | Iter.6 | Iter.7 | Iter.8 | Iter.9 | Iter.10 | Avg. |
|---|---|---|---|---|---|---|---|---|---|---|---|
| Full-KV | 4.39 | 4.72 | 4.18 | 4.63 | 4.29 | 4.51 | 4.37 | 4.89 | 4.11 | 4.46 | 4.45 |
| DuoAttention | 5.18 | 5.31 | 5.15 | 5.27 | 5.19 | 5.24 | 5.21 | 5.28 | 5.20 | 5.27 | 5.23 |
| SnapKV | 4.99 | 5.32 | 4.79 | 5.11 | 5.46 | 4.93 | 5.28 | 4.67 | 5.19 | 5.43 | 5.07 |
| PyramidKV | 4.78 | 5.34 | 4.91 | 5.26 | 4.97 | 5.18 | 4.85 | 5.11 | 4.93 | 5.69 | 5.02 |
| HeadKV-R | 4.73 | 4.95 | 4.81 | 4.92 | 4.68 | 4.90 | 4.79 | 4.99 | 4.84 | 4.88 | 4.86 |
| HeadKV-R2 | 4.91 | 4.35 | 4.66 | 4.82 | 4.58 | 4.79 | 4.47 | 4.99 | 4.23 | 4.64 | 4.72 |
| REAL | 4.55 | 4.71 | 4.48 | 4.66 | 4.60 | 4.74 | 4.52 | 4.69 | 4.57 | 4.63 | 4.63 |

Table 5: Latencies(/s) across 10 iterations for first-token generation on Mistral-7B-Instruct (Jiang et al., 2023). REAL achieves nearly identical latency to Full-KV and consistently outperforms existing compression baselines, showing the success-failure aware effectiveness with minimal overhead.

| Model | Iter.1 | Iter.2 | Iter.3 | Iter.4 | Iter.5 | Avg. |
|---|---|---|---|---|---|---|
| Llama-3-8B-Instruct | 0.0247 | 0.0272 | 0.0258 | 0.0262 | 0.0263 | 0.02604 |
| Mistral-7B-Instruct-v0.2 | 0.0270 | 0.0286 | 0.0267 | 0.0282 | 0.0288 | 0.02786 |

Table 6: Latencies(/s) to build the attention behavior matrix for different models across 5 iterations. The marginal latencies (approximately 26-28ms) demonstrate REAL's high efficiency. This one-time profiling cost is negligible compared to the overall inference time of long-context tasks.

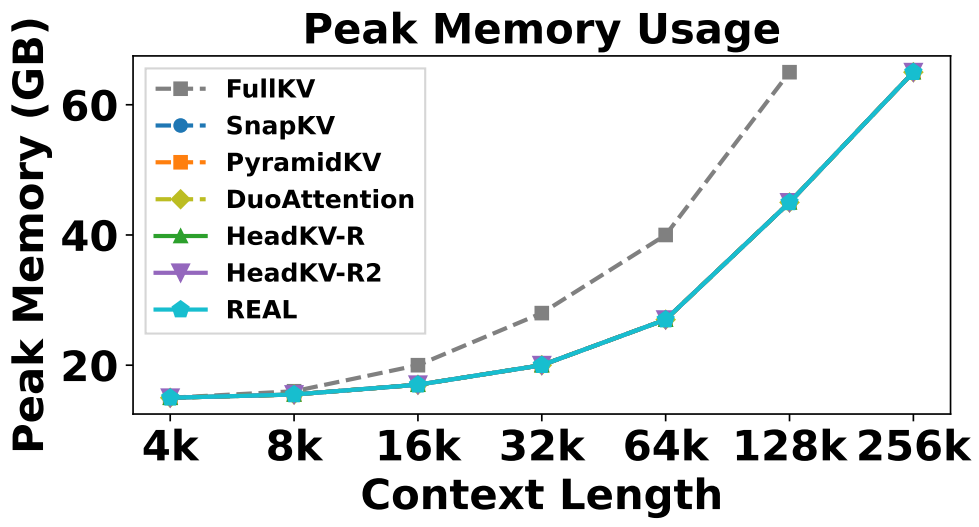

Figure 12: Peak memory comparison between REAL and baseline across context lengths. REAL achieves comparable memory efficiency with baselines while maintaining superior accuracy, demonstrating that attention behavior-based KV allocation does not incur additional memory overhead.

indicating that REAL is suitable for real-time deployment scenarios.

**Peak memory.** We evaluate peak memory usage with a maximum sequence length of 32K averaged over ten runs, as shown in Figure 12. Compared with Full-KV, REAL significantly reduces memory usage to a level comparable to existing baselines while maintaining superior accuracy, proving that success-failure aware allocation enables efficient compression.

## 6 Conclusion

In this paper, we propose ***RE**trieval-re**A**soning and **L**ogic-constructed (REAL)*, a novel head-wise KV cache compression method that addresses the limitations of existing approaches by recognizing and leveraging the functional heterogeneity of attention heads. We identify four comprehensive and distinct attention behaviors: retrieval-reasoning, bias, distraction, and widespread. We introduce *INFerence score (INFsc)*, defined as the harmonic mean of *Retrieval-reAsoning score (RAsc)* and *Logic-Constructed score (LCsc)*, which serves as a principled metric to guide KV budget allocation.

Extensive evaluations on diverse tasks from LongBench (Bai et al., 2024) and LongBench v2 (Bai et al., 2025) demonstrate that REAL consistently outperforms state-of-the-art baselines across a wide range of context lengths and model architectures. Moreover, REAL achieves substantial reductions in both TTFT and peak memory usage, making it particularly suitable for resource-constrained environments and real-world deployment scenarios. These results demonstrate that a fine-grained allocation strategy can effectively balance the trade-off between cache ratio and model performance, paving the way for more efficient long-context language models.

## Limitations

We acknowledge a few limitations of our work. While REAL demonstrates strong performance on English benchmarks, its effectiveness in multilingual and cross-lingual scenarios remains underexplored. The attention behaviors may not generalize to languages with different linguistic structures. Future work should investigate whether the proposed *INFsc* metric and head-wise allocation mechanism remain effective across diverse languages and whether language-specific adaptations are necessary for optimal performance.

## Acknowledgments

We appreciate the constructive comments provided by all anonymous reviewers. This work was supported by the National Research Foundation (NRF) of Korea under Grant RS-2025-00560896.

# A Appendix

## A.1 The Use of LLMs

We used LLMs to assist with grammar checking and correction. All ideas and technical content were developed entirely by the authors.

## A.2 Related Work

### A.2.1 KV Cache Eviction

KV cache plays a critical role in improving inference efficiency for both large language models (LLMs) and multimodal large language models (MLLMs) (Zeng et al., 2026; Zhang et al., 2026). Early representative methods, such as H2O (Zhang et al., 2023), SnapKV (Li et al., 2024), and Scissorhands (Liu et al., 2023), typically impose a uniform, fixed budget across all attention heads. This rigid design often leads to a significant degradation in generation quality.

Subsequent research has shifted toward importance-based strategies, such as HeadKV (Fu et al., 2025), DuoAttention (Xiao et al., 2025), and RazorAttention (Tang et al., 2025), which operate on the core assumption that only a small subset of cache entries remains consistently critical. However, these methods are primarily confined to a success-oriented paradigm. They analyze head behaviors only in successful retrieval-reasoning cases, while overlooking the diverse behaviors that arise in failure scenarios, such as bias and distraction. This oversight limits the potential to fully leverage the functional heterogeneity across attention heads for performance optimization.

Inspired by the confusion matrix, we reveal that head behaviors can be categorized into four distinct quadrants. Specifically, biased attention (false positives) and distracted attention (false negatives) are the primary culprits for degraded inference quality. By maximizing the signal-to-noise ratio (Shannon, 1948) — strengthening valid reasoning pathways while inhibiting noise — REAL introduces the first failure-aware budget allocation strategy. REAL transforms KV cache eviction from a greedy search into a strategic game between success and failure, pioneering a shift toward robust, multi-behavior-aware long-context modeling.

### A.2.2 Sparse Attention

In extensive contexts, not all tokens are equally important. Only a small subset of tokens drives generation (Li et al., 2025; Jiang et al., 2024; Liu et al., 2025). Exploiting this inherent sparsity is crucial for breaking the physical barriers of LLM inference: memory capacity and latency.

Two complementary approaches tackle these challenges. Eviction alleviates memory constraints by pruning non-essential history to fit physical limits, while Sparse Attention reduces compute overhead by dynamically filtering for key tokens. By exploiting sparsity across space and time dimensions, respectively, these two mechanisms form an orthogonal and decoupled relationship in resource management. REAL enables a dual-strategy approach, paving the way for long-context inference that is both memory-lightweight and latency-efficient.

### A.2.3 LLM Quantization

LLM quantization can be largely classified into weight-only and weight-activation.

**Weight-only quantization (WOQ).** It was introduced to address memory bottlenecks and has enabled low-bit compression (Frantar et al., 2022; Dettmers et al., 2024; Lee et al., 2023; Lin et al., 2025; Dettmers et al., 2023; Chee et al., 2023). However, in serving scenarios, batching amortizes memory overhead and shifts the bottleneck from memory-bound to compute-bound. Consequently, WOQ fails to utilize efficient low-bit hardware instructions, limiting its ability to deliver meaningful throughput gains.

**Weight-activation quantization (WAQ).** Mainstream strategies for handling activation outliers, including Mixed Precision (Dettmers et al., 2022), Math Transform (Xiao et al., 2023; Shao et al., 2024; Yao et al., 2022; Wei et al., 2023), Reordering (Yuan et al., 2023), Low-Rank (Liu et al., 2024; Wu et al., 2023), and Co-design (Guo et al., 2023; Zhou et al., 2025), all exhibit significant limitations. They either suffer severe accuracy degradation at ultra-low bit widths or incur high latency due to computational complexity and hardware constraints. Since KV is a subset of activations, it inherits these intractable outlier challenges. In contrast, our proposed REAL method breaks this tradeoff, simultaneously achieving extreme acceleration and lossless accuracy.

## A.3 Three Score Analysis

Figure 13 presents the score ranking of layer-heads in descending order.

## A.4 Dataset Details

Table 7 shows the details of sixteen QA datasets from LongBench (Bai et al., 2024) and eleven

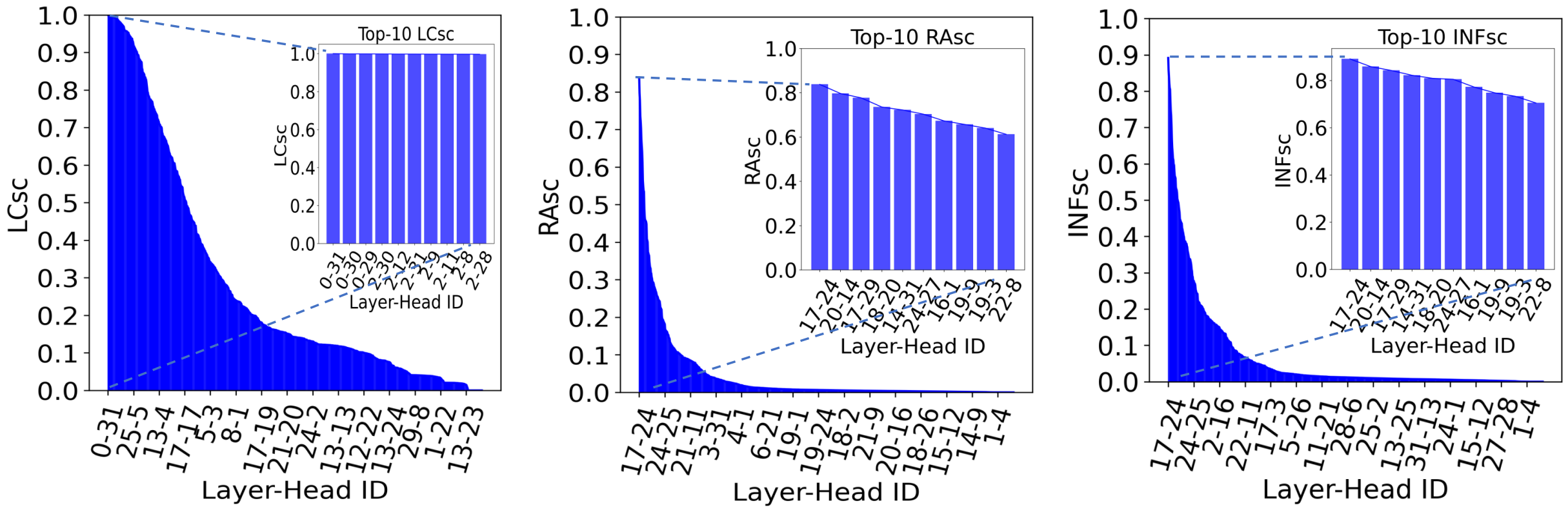

Figure 13: Layer-heads ranked by descending scores on Llama-3-8B-Instruct (Grattafiori et al., 2024), each with a zoomed-in bar chart for the top-10 layer-head IDs.

| Label | Task | Avg. |
|---|---|---|
| NrtvQA | NarrativeQA | 18,409 |
| Qasper | Qasper | 3,619 |
| MF-en | MultiFieldQA-EN | 4,559 |
| HotpotQA | HotpotQA | 9,151 |
| 2Wiki | 2WikiMultiHopQA | 4,887 |
| Musique | Musique | 11,214 |
| QMSum | QMSum | 10,614 |
| MultiNews | MultiNews | 2,113 |
| TREC | TREC | 5,177 |
| TriviaQA | TriviaQA | 8,209 |
| SAMSum | SAMSum | 6,258 |
| PCount | PassageCount | 11,141 |
| PRe | PassageRetrieval-en | 9,289 |
| LCC | LCC | 1,235 |
| RB-P | RepoBench-P | 4,206 |
| GovReport | Government report | 8,734 |
| Literary | Literary | 72K |
| Legal | Legal | 28K |
| Detective | Detective | 70K |
| Academic | Academic | 27K |
| Financial | Financial | 49K |
| Govern | Government reports | 20K |
| UserGuide | User guide QA | 61K |
| Many-Shot | Many-Shot | 71K |
| Dialogue | Dialogue history QA | 77K |
| Table | Table QA | 42K |
| KnGraph | Knowledge graph reasoning | 52K |

Table 7: Dataset details.

extended-length datasets from LongBench v2 (Bai et al., 2025).

### A.5 Pareto Analysis

REAL demonstrates a highly efficient Pareto-optimal front. Evaluation on LongBench (Bai et al., 2024) with Llama-3-8B-Instruct (Grattafiori et al., 2024) reveals two primary insights into the tradeoffs between resource consumption and performance. For the memory-accuracy tradeoff in Table 8, REAL maintains highly competitive accuracy while achieving nearly an $8\times$ reduction in memory footprint compared to Full-KV. Regarding the latency-accuracy tradeoff in Table 9, REAL consistently delivers approximately a $2\times$ decoding speedup. To reflect the empirical generation characteristics of HotpotQA (832.6 tokens) and 2WikiMultihopQA (325.4 tokens), decoding latency is benchmarked across sequence lengths of 256, 512, and 1024, averaged over five iterations.

| KV Budget (Memory) | Accuracy |
|---|---|
| 64 (56M) | 38.26 |
| 128 (108M) | 40.43 |
| 256 (210M) | 40.96 |
| 512 (439M) | 41.13 |
| 1024 (878M) | 41.26 |
| Full-KV (6870M) | 41.86 |

Table 8: Memory-accuracy trade-off of REAL.

| Generation Length | REAL (s) | FullKV (s) |
|---|---|---|
| 256 | 11.82 | 21.33 |
| 512 | 17.06 | 39.52 |
| 1024 | 33.64 | 75.91 |

Table 9: Decoding latency when KV Budget = 256.

### A.6 Synthetic Needles Analysis

#### A.6.1 Retrieval-Reasoning Difficulty

Table 10 presents other representative Question-Needle pairs to demonstrate the model's multi-dimensional information retrieval capabilities in complex long-context environments. As measured by the OpenAI GPT-5.x and O1/O3 tokenizers, Table 11 presents the token counts for different methods. Notably, the proportion of answer-related tokens in these cases (averaging 18%) is significantly lower than that observed in retrieval (68.97%) and retrieval-reasoning (52.17%) methods. This discrepancy suggests that, while retrieval-based methods rely heavily on direct keyword matching, the observed attention behaviors incorporate a broader context, potentially capturing more complex structural logic.

| Example |
|---|
| **3. Question:** How many eggs did the chef buy?<br>**Needle:** Regarding nutritional profiles, culinary experts praise eggs as a powerhouse ingredient, providing roughly six grams of high-quality protein and essential vitamins per serving. For the morning special, the chef went to the market and bought half a dozen fresh eggs. The chef set the timer because the eggs needed exactly three minutes to fry. The chef walked over to the large stainless steel fridge and took out eight eggs to prepare the batter. |
| **4. Question:** What is the main export of the city located in Europe?<br>**Needle:** Renowned for its romantic art and historic vineyards, France is the proud home of City Alpha, while Japan, a global leader in advanced technology and robotics, hosts City Beta. The main export of City Alpha is fine wine and cheese. The main export of City Beta is electronics and cars. The main exports of City Gamma are coffee beans and textiles. |

Table 10: Other manual needles.

| Method | Needle | Answer-related | Bias | Distraction |
|---|---|---|---|---|
| Retrieval (HeadKV-R) | 29 | 20 (68.97%) | N/A | N/A |
| Retrieval-Reasoning (HeadKV-R2) | 69 | 36 (52.17%) | N/A | N/A |
| Example 1 (Figure 3) | 103 | 15 (14.56%) | 15 (14.56%) | 7 (6.80%) |
| Example 2 | 127 | 23 (18.11%) | 25 (19.69%) | 19 (14.96%) |
| Example 3 | 85 | 11 (12.94%) | 15 (17.65%) | 24 (28.24%) |
| Example 4 | 72 | 19 (26.39%) | 16 (22.22%) | 12 (16.67%) |

Table 11: Detailed analysis of token counts across different attention behaviors.

| Metric | Direction | Description and Significance |
|---|---|---|
| Pearson $r$ | High (↑) | Measures the degree of linear correlation between variables. |
| Spearman $\rho$ | High (↑) | Evaluates the monotonic relationship based on rank correlation. |
| RMSE | Low (↓) | Root mean square error quantifies absolute prediction deviation. |
| $p$-value | Significance | Statistical reliability. $p < 0.01$ ensures the correlation is not due to random chance. |

Table 12: Definitions and descriptions of correlation metrics.

| Strategy | Pearson r (↑) | Spearman $\rho$ (↑) | RMSE (↓) |
|---|---|---|---|
| Random | 0.0138 ($p = 0.66$) | -0.0159 ($p = 0.61$) | 0.17 |
| Synthetic | 0.8671 ($p < 0.01$) | 0.9004 ($p < 0.01$) | 0.06 |

Table 13: Correlation analysis for narrativeQA (Single-Doc, 18.4K Tokens) (Bai et al., 2024).

| Strategy | Pearson r (↑) | Spearman $\rho$ (↑) | RMSE (↓) |
|---|---|---|---|
| Random | -0.0090 ($p = 0.77$) | -0.0183 ($p = 0.56$) | 0.37 |
| Synthetic | 0.7823 ($p < 0.01$) | 0.8628 ($p < 0.01$) | 0.27 |

Table 14: Correlation analysis for 2WikiMultihopQA (Multi-Doc, 4.9K Tokens) (Bai et al., 2024).

### A.6.2 Generalizable Categories

We investigate the intrinsic relationship between synthetic needles and task-extracted needles. According to Table 12, the quantitative correlation analysis in Tables 13 and 14 yields three primary observations regarding the effectiveness of the synthetic strategy. First, the synthetic approach exhibits a strong statistical correlation with task-extracted needles, achieving Pearson's $r$ and Spearman's $\rho$ of up to 0.90 ($p < 0.01$). Second, it demonstrates minimal structural deviation, as indicated by the lowest RMSE across all evaluated datasets. Finally, the synthetic strategy offers a substantial reduction in computational overhead. While task-based extraction requires several hours of processing, the synthetic method completes in seconds on a dual-RTX 4090 GPU configuration, delivering a significant throughput improvement for large-scale inference profiling.

## A.7 Comprehensive Results

| Token Budget | Method | narrativeqa | qasper | multi-fieldqa | hotpotqa | 2wikimqa | musique | gov-report | qmsum | multi-news | trec | triviaqa | samsum | passagecnt | pass-ret. | lcc | repobench-p | Avg. |
|---|---|---|---|---|---|---|---|---|---|---|---|---|---|---|---|---|---|---|
| N/A | Full-KV | 25.56 | 32.07 | 39.71 | 43.57 | 35.28 | 21.18 | 28.71 | 23.26 | 26.64 | 73.50 | 90.48 | 42.33 | 4.80 | 69.25 | 59.29 | 54.05 | 41.86 |
| 64 | DuoAttention | 7.92 | 7.04 | 13.56 | 18.23 | 12.84 | 8.42 | 8.47 | 8.70 | 8.74 | 28.47 | 39.64 | 16.91 | 2.44 | 29.80 | 26.06 | 39.62 | 17.31 |
| | SnapKV | 10.01 | 6.34 | 18.77 | 20.06 | 13.80 | 8.94 | 9.72 | 10.24 | 10.20 | 32.41 | 44.85 | 19.23 | 2.84 | 34.00 | 29.46 | 38.82 | 19.36 |
| | PyramidKV | 13.13 | 8.70 | 19.45 | 21.88 | 14.18 | 9.69 | 10.86 | 11.60 | 11.50 | 36.15 | 49.88 | 21.41 | 3.15 | 37.97 | 32.72 | 39.14 | 21.34 |
| | HeadKV-R | 15.61 | 19.17 | 28.84 | 26.09 | 21.63 | 13.56 | 15.59 | 16.84 | 16.63 | 51.84 | 71.34 | 30.66 | 4.48 | 54.52 | 46.75 | 51.05 | 30.29 |
| | HeadKV-R2 | 17.59 | 21.88 | 31.79 | 31.56 | 24.85 | 15.03 | 17.84 | 19.37 | 19.10 | 59.04 | 81.12 | 34.94 | 5.19 | 62.11 | 53.16 | 55.53 | 34.38 |
| | REAL | **22.02** | **23.72** | **32.88** | **37.11** | **26.21** | **17.52** | **19.93** | **21.73** | **21.40** | **65.96** | **90.53** | **39.01** | **5.78** | **69.42** | **59.31** | **59.67** | **38.26** |
| 128 | DuoAttention | 10.39 | 8.58 | 16.59 | 22.20 | 15.68 | 10.33 | 10.45 | 10.98 | 10.95 | 34.87 | 48.29 | 20.69 | 3.04 | 36.57 | 31.71 | 42.22 | 20.85 |
| | SnapKV | 15.63 | 11.81 | 23.03 | 30.64 | 21.72 | 14.37 | 14.65 | 15.82 | 15.63 | 48.52 | 66.74 | 28.73 | 4.27 | 51.02 | 43.76 | 47.74 | 28.38 |
| | PyramidKV | 14.94 | 12.78 | 22.78 | 27.73 | 20.80 | 12.95 | 14.55 | 13.55 | 15.06 | 47.72 | 64.12 | 26.89 | 3.60 | 49.53 | 43.01 | 50.98 | 27.56 |
| | HeadKV-R | 19.38 | 22.83 | 33.11 | 35.86 | 28.11 | 16.77 | 17.78 | 18.55 | 18.83 | 60.87 | 75.09 | 32.12 | 4.00 | 58.24 | 51.50 | 55.59 | 34.29 |
| | HeadKV-R2 | 19.21 | 27.60 | **38.70** | 39.53 | 32.02 | 18.41 | 19.69 | 19.77 | 21.63 | 64.69 | 80.98 | 34.88 | **6.02** | 62.29 | 55.08 | 57.48 | 37.38 |
| | REAL | **24.69** | **29.49** | 37.04 | **43.36** | **33.37** | **20.07** | **22.73** | **22.47** | **24.07** | **73.12** | **87.62** | **39.13** | 5.10 | **67.21** | **58.47** | **58.89** | **40.43** |
| 256 | DuoAttention | 14.04 | 10.84 | 21.07 | 28.07 | 19.88 | 13.15 | 13.38 | 14.35 | 14.21 | 44.35 | 61.09 | 26.28 | 3.91 | 46.60 | 40.08 | 46.06 | 26.09 |
| | SnapKV | 17.06 | 16.29 | 29.87 | 32.59 | 25.94 | 15.15 | 17.11 | 18.54 | 18.29 | 56.69 | 77.92 | 33.55 | 4.97 | 59.62 | 51.08 | 54.34 | 33.06 |
| | PyramidKV | 18.00 | 16.64 | 28.95 | 32.89 | 24.58 | 15.43 | 17.07 | 18.50 | 18.26 | 56.43 | 77.55 | 33.42 | 4.99 | 59.36 | 50.84 | 53.76 | 32.91 |
| | HeadKV-R | 19.27 | 26.24 | **34.29** | 35.00 | 28.32 | 17.06 | 20.02 | 21.81 | 21.48 | 66.26 | 90.97 | 39.19 | 5.81 | 69.74 | 59.59 | 60.33 | 38.46 |
| | HeadKV-R2 | 19.93 | **27.59** | 32.73 | 36.61 | 30.92 | 16.90 | 20.60 | 22.43 | 22.10 | 68.04 | 93.38 | 40.27 | 6.02 | 71.60 | 61.20 | 62.10 | 39.52 |
| | REAL | **21.67** | 27.53 | 31.68 | **37.45** | **33.41** | **18.99** | **21.34** | **23.27** | **22.91** | **70.62** | **96.91** | **41.77** | **6.20** | **74.32** | **63.50** | **63.83** | **40.96** |
| 512 | DuoAttention | 17.08 | 12.70 | 24.81 | 32.98 | 23.39 | 15.48 | 15.80 | 17.15 | 16.92 | 52.32 | 71.88 | 30.96 | 4.60 | 55.04 | 47.11 | 49.26 | 30.47 |
| | SnapKV | 20.29 | 20.60 | 32.24 | 35.22 | 27.98 | 15.87 | 19.06 | 20.71 | 20.41 | 62.82 | 86.25 | 37.21 | 5.61 | 66.10 | 56.56 | 58.31 | 36.57 |
| | PyramidKV | 21.22 | 21.53 | 30.82 | 37.62 | 27.15 | 17.40 | 19.48 | 21.26 | 20.93 | 64.34 | 88.28 | 38.07 | 5.68 | 67.73 | 57.84 | 57.57 | 37.30 |
| | HeadKV-R | 19.21 | 26.36 | **34.09** | 36.93 | 30.75 | 17.03 | 20.55 | 22.39 | 22.05 | 68.00 | 93.34 | 40.22 | 5.96 | 71.56 | 61.15 | 61.86 | 39.46 |
| | HeadKV-R2 | 20.49 | 27.13 | 32.78 | **37.54** | 31.53 | 18.39 | 20.98 | 22.88 | 22.54 | 69.43 | 95.28 | 41.06 | 6.09 | 73.07 | 62.42 | 62.64 | 40.26 |
| | REAL | **21.57** | **28.76** | 33.33 | 36.88 | **31.50** | **19.30** | **21.42** | **23.35** | **22.99** | **70.92** | **97.36** | **41.93** | **6.19** | **74.64** | **63.77** | **64.18** | **41.13** |
| 1024 | DuoAttention | 18.78 | 13.76 | 26.90 | 35.71 | 25.35 | 16.80 | 17.18 | 18.73 | 18.44 | 56.71 | 77.81 | 33.56 | 5.02 | 59.69 | 50.99 | 51.06 | 32.91 |
| | SnapKV | 21.80 | 25.10 | 33.49 | 36.96 | 30.23 | 17.04 | 20.60 | 22.46 | 22.12 | 67.97 | 93.26 | 40.23 | 6.04 | 71.54 | 61.13 | 61.20 | 39.44 |
| | PyramidKV | **21.16** | 24.23 | 31.71 | **37.28** | 29.37 | 18.24 | 20.26 | 22.08 | 21.75 | 67.01 | 91.96 | 39.63 | 5.88 | 70.53 | 60.24 | 60.57 | 38.86 |
| | HeadKV-R | 20.40 | 28.22 | **34.33** | 36.30 | 31.44 | 16.96 | 20.97 | 22.86 | 22.52 | 69.29 | 95.07 | 41.00 | 6.12 | 72.91 | 62.32 | 62.79 | 40.21 |
| | HeadKV-R2 | 20.32 | 28.16 | 34.22 | 36.95 | 31.46 | **18.91** | 21.27 | 23.18 | 22.83 | 70.30 | 96.48 | 41.60 | 6.18 | 74.00 | 63.23 | 63.48 | 40.78 |
| | REAL | 20.94 | **29.52** | 33.80 | 37.24 | **31.53** | 18.89 | **21.50** | **23.43** | **23.07** | **71.12** | **97.62** | **42.06** | **6.24** | **74.85** | **63.96** | **64.46** | **41.26** |

Table 15: Question-aware, individual results of LongBench (Bai et al., 2024) for Llama-3-8B-Instruct (Grattafiori et al., 2024).

| Token Budget | Method | narrativeqa | qasper | multi-fieldqa | hotpotqa | 2wikimqa | musique | gov-report | qmsum | multi-news | trec | triviaqa | samsum | passagecnt | pass-ret. | lcc | repobench-p | Avg. |
|---|---|---|---|---|---|---|---|---|---|---|---|---|---|---|---|---|---|---|
| N/A | Full-KV | 26.63 | 32.99 | 49.34 | 42.77 | 27.35 | 18.78 | 32.87 | 23.24 | 27.10 | 71.00 | 86.23 | 42.79 | 2.75 | 86.98 | 56.93 | 54.49 | 42.64 |
| 64 | SnapKV | 9.59 | 8.27 | 27.80 | 20.88 | 11.03 | 3.45 | 8.37 | 10.14 | 8.93 | 45.83 | 62.12 | 25.92 | 1.30 | 48.33 | 36.73 | 32.05 | 22.55 |
| | PyramidKV | 10.29 | 8.99 | 27.43 | 21.56 | 12.25 | 4.64 | 9.52 | 10.40 | 10.06 | 50.27 | 67.19 | 28.05 | 1.35 | 55.96 | 40.19 | 35.87 | 24.62 |
| | DuoAttention | 15.09 | 13.49 | 26.44 | 30.50 | 22.22 | 13.17 | 16.53 | 15.96 | 16.44 | 41.18 | 48.10 | 24.52 | 1.41 | 41.18 | 30.79 | 29.07 | 24.13 |
| | HeadKV-R | 14.27 | 15.26 | 36.06 | 28.86 | 16.09 | 7.45 | 10.32 | 11.92 | 11.64 | 52.56 | 67.84 | 29.44 | 1.45 | 61.34 | 42.30 | 38.75 | 27.85 |
| | HeadKV-R2 | 16.40 | 21.20 | 43.02 | 34.75 | 21.39 | 10.84 | 17.54 | 16.76 | 17.42 | 59.90 | 74.08 | 33.02 | 1.55 | 63.10 | 46.94 | 43.07 | 32.56 |
| | REAL | **21.34** | **24.45** | **44.31** | **37.04** | **22.14** | **14.34** | **23.51** | **19.20** | **21.75** | **65.04** | **77.27** | **36.98** | **1.87** | **75.48** | **48.02** | **45.05** | **36.11** |
| 128 | SnapKV | 10.96 | 11.44 | 34.73 | 23.37 | 11.32 | 5.02 | 10.25 | 12.21 | 10.87 | 52.23 | 70.29 | 29.71 | 1.51 | 54.55 | 41.69 | 36.50 | 26.04 |
| | PyramidKV | 14.00 | 14.22 | 35.96 | 25.00 | 14.97 | 7.83 | 13.47 | 14.40 | 19.04 | 58.05 | 75.89 | 33.01 | 1.58 | 62.44 | 45.81 | 36.26 | 29.50 |
| | DuoAttention | 20.21 | 18.59 | 34.15 | 35.26 | **27.04** | **18.01** | 21.62 | 20.99 | 21.53 | 50.81 | 58.54 | 30.54 | 1.42 | 49.14 | 37.55 | 35.63 | 30.06 |
| | HeadKV-R | 15.92 | 21.55 | 40.35 | 31.61 | 18.26 | 10.08 | 14.14 | 14.81 | 14.52 | 58.43 | 74.39 | 33.11 | 1.56 | 66.44 | 46.55 | 42.84 | 31.53 |
| | HeadKV-R2 | 18.22 | 21.13 | 41.66 | 34.46 | 20.83 | 11.23 | 17.48 | 16.66 | 17.36 | 61.77 | 76.81 | 33.90 | 1.83 | 65.81 | 48.67 | 44.57 | 33.27 |
| | REAL | **23.99** | **28.22** | **46.45** | **38.78** | 26.34 | 16.38 | **26.23** | **21.64** | **24.36** | **71.03** | **84.05** | **40.58** | **2.06** | **81.58** | **52.34** | **49.17** | **39.58** |
| 256 | SnapKV | 13.74 | 16.42 | 39.78 | 28.24 | 16.64 | 6.41 | 13.98 | 18.88 | 17.37 | 45.39 | 75.85 | 26.69 | 2.07 | 54.80 | 40.23 | 37.75 | 28.39 |
| | PyramidKV | 16.06 | 20.00 | 42.58 | 33.39 | 20.46 | 9.94 | 17.43 | 22.39 | 20.86 | 50.60 | 81.71 | 30.36 | 2.12 | 58.78 | 44.08 | 41.63 | 32.02 |
| | DuoAttention | 22.25 | 21.00 | 37.76 | 38.78 | **29.39** | 18.57 | 21.91 | 23.70 | 22.86 | 50.49 | **86.24** | 36.99 | 2.26 | 56.41 | **54.70** | 43.44 | 36.05 |
| | HeadKV-R | 19.09 | 23.42 | 43.12 | 35.47 | 21.27 | 11.45 | 19.15 | 24.61 | 22.93 | 53.48 | 79.65 | 38.86 | 2.53 | 64.70 | 43.94 | 56.74 | 35.03 |
| | HeadKV-R2 | 19.09 | 25.87 | 45.61 | 36.96 | 20.99 | 13.32 | 20.37 | **25.93** | 24.21 | 55.37 | 82.60 | 40.46 | 2.62 | 66.80 | 45.81 | 58.67 | 36.54 |
| | REAL | **25.78** | **29.40** | **47.87** | **41.12** | 26.16 | **17.84** | **24.93** | 25.62 | **28.87** | **61.39** | 85.00 | **45.50** | **2.73** | **84.31** | 51.23 | **64.16** | **41.37** |
| 512 | SnapKV | 16.30 | 20.99 | 40.86 | 30.96 | 17.60 | 7.16 | 16.51 | 19.02 | 18.77 | 48.99 | 74.73 | 35.39 | 2.53 | 57.63 | 48.01 | 49.40 | 31.55 |
| | PyramidKV | 20.84 | 26.74 | 46.14 | 37.31 | 23.40 | 14.36 | 22.47 | 25.01 | 24.76 | 58.13 | 84.31 | 41.59 | 2.56 | 69.26 | 54.38 | 50.78 | 37.63 |
| | DuoAttention | 25.22 | 24.66 | 41.63 | 38.05 | **31.37** | **20.85** | **26.47** | 25.76 | 22.64 | 61.83 | 82.65 | 38.46 | 2.53 | 57.56 | 53.38 | 53.86 | 37.93 |
| | HeadKV-R | 21.66 | 27.63 | 46.14 | 38.94 | 23.03 | 15.23 | 23.00 | 25.71 | 25.44 | 59.83 | 88.38 | 43.37 | 2.73 | 68.23 | 56.99 | 58.49 | 39.05 |
| | HeadKV-R2 | 21.78 | 27.52 | 46.45 | 38.85 | 23.39 | 15.56 | 23.13 | 25.91 | 25.63 | 60.38 | 89.85 | 43.99 | 2.80 | 69.71 | 57.93 | 59.47 | 39.52 |
| | REAL | **25.72** | **28.72** | **47.08** | **40.21** | 25.91 | 16.99 | 25.07 | **27.87** | **27.59** | **63.44** | **93.32** | **46.14** | **2.82** | **72.24** | **60.23** | **61.78** | **41.57** |
| 1024 | SnapKV | 16.49 | 21.33 | 40.40 | 32.95 | 16.21 | 9.19 | 16.13 | 17.90 | 17.16 | 49.12 | 76.40 | 35.71 | 2.70 | 74.40 | 46.58 | 50.24 | 32.68 |
| | PyramidKV | 23.88 | 29.65 | 48.77 | 40.09 | 26.03 | 18.40 | 24.41 | 26.17 | 25.43 | 60.54 | 87.59 | 43.83 | 2.77 | 82.19 | 54.61 | 58.23 | 40.79 |
| | DuoAttention | 25.77 | 25.75 | 42.34 | 37.49 | 29.32 | 19.47 | **26.11** | 24.51 | 25.23 | 60.08 | 89.82 | 40.56 | 3.17 | 83.30 | 55.25 | 54.96 | 40.20 |
| | HeadKV-R | 24.45 | **31.35** | **49.30** | 40.61 | 26.69 | 17.70 | 24.68 | 26.51 | 25.74 | 62.34 | 90.55 | 44.92 | 3.12 | 83.92 | 56.16 | 59.94 | 41.75 |
| | HeadKV-R2 | **26.84** | 26.82 | 43.41 | 40.56 | **30.39** | **20.54** | 24.18 | 25.58 | **26.30** | 61.69 | 86.43 | 41.63 | 3.17 | 84.37 | 54.32 | **66.03** | 42.01 |
| | REAL | 25.77 | 30.99 | 47.98 | **41.61** | 27.04 | 18.35 | 25.15 | **27.00** | 26.22 | **63.22** | **91.72** | **45.61** | **3.34** | **87.04** | **56.97** | 60.79 | **42.30** |

Table 16: Question-aware, individual results of LongBench (Bai et al., 2024) for Mistral-7B-Ins-v0.2 (Jiang et al., 2023).

| Token Budget | Method | Literary | Legal | Detective | Academic | Financial | Govern | UserGuide | Many-shot | DialogueHist | Table | KnGraph | Avg. |
|---|---|---|---|---|---|---|---|---|---|---|---|---|---|
| Full | Full-KV | 24.34 | 47.89 | 33.28 | 19.40 | 21.42 | 11.34 | 34.98 | 27.04 | 46.28 | 31.70 | 46.00 | 31.24 |
| 64 | DuoAttention | 5.46 | 13.12 | 8.61 | 3.83 | 4.03 | 1.68 | 8.86 | 6.23 | 12.85 | 7.98 | 12.21 | 7.71 |
| | SnapKV | 7.91 | 15.90 | 10.92 | 6.50 | 6.92 | 3.83 | 11.40 | 8.95 | 15.23 | 10.59 | 15.04 | 10.29 |
| | PyramidKV | 7.59 | 15.56 | 10.25 | 6.11 | 6.88 | 3.44 | **11.47** | 8.38 | 14.90 | 9.67 | 14.63 | 9.90 |
| | HeadKV-R | 8.08 | 16.52 | 10.92 | **6.81** | 7.18 | 3.95 | 11.37 | 9.52 | 15.64 | 10.84 | 16.10 | 10.63 |
| | HeadKV-R2 | 4.80 | 19.91 | 8.58 | 3.57 | **10.54** | 6.52 | 9.62 | **15.09** | 18.70 | 10.91 | 21.39 | 11.79 |
| | REAL | **8.68** | **23.12** | **14.03** | 6.78 | 6.94 | **6.91** | 9.02 | 10.39 | **22.00** | **13.27** | **21.74** | **12.90** |
| 128 | DuoAttention | 5.68 | 13.55 | 8.73 | 4.04 | 4.73 | 1.04 | 8.83 | 6.47 | 13.38 | 8.26 | 12.53 | 7.93 |
| | SnapKV | 8.15 | 16.38 | 10.92 | 6.69 | 7.11 | 3.95 | 11.37 | 9.19 | 15.64 | 10.87 | 15.41 | 10.52 |
| | PyramidKV | 9.13 | 18.77 | 12.24 | 7.47 | 7.97 | 4.48 | 12.77 | 10.49 | 17.49 | 12.36 | 17.57 | 11.89 |
| | HeadKV-R | 9.61 | 19.73 | 12.91 | 7.86 | 8.40 | 4.71 | 13.47 | 11.03 | 18.42 | 13.00 | 18.49 | 12.51 |
| | HeadKV-R2 | 14.17 | 28.49 | 19.14 | 11.48 | 12.38 | 6.78 | 20.01 | 16.09 | 27.07 | 18.77 | 26.77 | 18.29 |
| | REAL | **15.87** | **32.80** | **21.37** | **13.10** | **14.09** | **7.85** | **22.32** | **18.44** | **30.78** | **22.03** | **29.67** | **20.76** |
| 256 | DuoAttention | 7.72 | 18.72 | 11.30 | 5.81 | 6.77 | 2.08 | 12.48 | 9.23 | 17.65 | 11.25 | 17.46 | 10.95 |
| | SnapKV | 10.59 | 21.93 | 14.24 | 8.96 | 9.53 | 5.16 | 15.11 | 12.57 | 20.46 | 14.52 | 20.15 | 13.93 |
| | PyramidKV | 11.00 | 22.27 | 14.81 | 8.96 | 9.81 | 5.31 | 15.67 | 12.44 | 21.06 | 14.74 | 20.33 | 14.22 |
| | HeadKV-R | 13.75 | 28.35 | 18.57 | 11.35 | 12.04 | 6.75 | 19.31 | 15.90 | 26.61 | 18.77 | 26.13 | 17.96 |
| | HeadKV-R2 | 14.48 | 29.79 | 19.57 | 11.93 | 12.89 | 7.12 | 20.71 | 16.71 | 28.00 | 19.81 | 26.91 | 18.90 |
| | REAL | **22.67** | **41.06** | **28.23** | **19.87** | **20.82** | **11.45** | **32.51** | **25.69** | **38.63** | **29.39** | **36.45** | **27.89** |
| 512 | DuoAttention | 8.87 | 22.31 | 11.90 | 7.65 | 7.26 | 4.13 | 12.07 | 11.54 | 19.44 | 13.62 | 18.20 | 12.45 |
| | SnapKV | 11.00 | 24.18 | 14.24 | 9.41 | 9.47 | 5.81 | 14.52 | 13.47 | 21.52 | 15.91 | 20.01 | 14.50 |
| | PyramidKV | 12.78 | 27.20 | 17.04 | 10.77 | 11.40 | 6.49 | 17.77 | 15.20 | 25.22 | 18.23 | 23.83 | 16.90 |
| | HeadKV-R | 12.76 | 35.11 | 20.02 | 8.72 | 9.88 | 10.72 | 23.30 | 14.47 | 23.41 | 20.27 | 32.41 | 19.19 |
| | HeadKV-R2 | 18.56 | 32.48 | 21.91 | 16.99 | 16.90 | 13.15 | 22.92 | 21.20 | 29.58 | 24.05 | 27.33 | 22.28 |
| | REAL | **23.11** | **43.84** | **29.56** | **19.81** | **20.96** | **13.58** | **31.11** | **26.39** | **41.12** | **30.56** | **39.05** | **29.01** |
| 1024 | DuoAttention | 9.91 | 22.75 | 14.30 | 7.23 | 8.42 | 3.10 | 15.28 | 11.21 | 21.63 | 13.85 | 21.00 | 13.47 |
| | SnapKV | 12.78 | 25.96 | 17.24 | 10.38 | 11.18 | 6.18 | 17.91 | 14.55 | 24.44 | 17.12 | 23.69 | 16.49 |
| | PyramidKV | 12.46 | 23.23 | 17.47 | 9.66 | 11.31 | 5.58 | 18.01 | 13.03 | 24.16 | 15.69 | 23.37 | 15.82 |
| | HeadKV-R | 19.05 | 36.41 | 27.89 | 14.72 | 15.74 | 8.38 | 27.45 | 21.74 | 39.88 | 23.84 | 37.93 | 24.82 |
| | HeadKV-R2 | 20.57 | 43.68 | 27.56 | 17.17 | 18.25 | 10.49 | 29.21 | 24.28 | 40.03 | 28.59 | 37.63 | 27.04 |
| | REAL | **22.51** | **45.59** | **30.56** | **18.33** | **19.96** | **10.94** | **31.76** | **25.63** | **42.67** | **30.50** | **42.09** | **29.14** |
| 2048 | DuoAttention | 3.21 | 27.98 | 20.34 | 9.59 | 11.30 | 3.68 | 21.63 | 14.38 | 27.70 | 17.96 | 30.59 | 18.03 |
| | SnapKV | 17.33 | 31.85 | 24.69 | 13.35 | 15.53 | 7.62 | 25.71 | 18.79 | 31.38 | 22.25 | 34.41 | 22.08 |
| | PyramidKV | 18.32 | 27.48 | 21.25 | 16.89 | 17.45 | **14.11** | 21.72 | 19.95 | 26.29 | 20.94 | 26.56 | 21.00 |
| | HeadKV-R | 22.03 | 41.09 | 26.79 | 17.89 | 17.89 | 10.00 | 30.36 | 24.74 | 38.97 | 25.61 | 42.69 | 27.10 |
| | HeadKV-R2 | 21.22 | 39.51 | 30.45 | 16.45 | 19.17 | 9.21 | 31.76 | 23.31 | 38.79 | **29.23** | 40.71 | 27.26 |
| | REAL | **23.07** | **43.82** | **32.02** | **18.00** | **20.46** | 10.26 | **32.81** | **26.17** | **43.60** | 28.91 | **44.07** | **29.38** |
| 4096 | DuoAttention | 15.93 | 31.41 | 23.35 | 12.17 | 13.96 | 6.28 | 24.41 | 17.67 | 30.69 | 21.01 | 33.70 | 20.96 |
| | SnapKV | 18.06 | 33.28 | 25.69 | 13.93 | 16.17 | 7.96 | 26.86 | 19.60 | 32.77 | 23.30 | 35.51 | 23.01 |
| | PyramidKV | 20.01 | 36.64 | 28.02 | 15.29 | 17.46 | 8.75 | 29.21 | 21.50 | 36.01 | 25.52 | **39.01** | 25.22 |
| | HeadKV-R | 20.25 | **42.86** | 29.02 | 15.62 | 19.66 | 9.70 | **32.26** | 22.31 | **41.10** | 27.42 | 38.87 | 27.19 |
| | HeadKV-R2 | 24.93 | 35.21 | 28.48 | 23.05 | 23.61 | 19.76 | 29.10 | 26.30 | 34.07 | 28.41 | 33.86 | 27.89 |
| | REAL | **26.81** | 37.78 | **30.42** | **25.21** | **25.45** | **21.28** | 31.31 | **28.47** | 36.81 | **30.80** | 36.42 | **30.07** |
| 8192 | DuoAttention | 19.68 | 28.75 | 22.81 | 18.28 | 18.69 | 15.41 | 23.39 | 20.91 | 27.86 | 22.19 | 27.98 | 22.36 |
| | SnapKV | 24.81 | 33.62 | 28.15 | 23.04 | 23.90 | **20.09** | 28.84 | 25.84 | 32.94 | 27.48 | 32.79 | 27.41 |
| | PyramidKV | 22.20 | **46.21** | 31.12 | 17.62 | 20.18 | 10.86 | 32.36 | 24.83 | **44.20** | 29.73 | 41.49 | 29.16 |
| | HeadKV-R | 25.05 | 38.00 | 29.24 | 22.99 | 23.75 | 18.95 | 30.21 | 27.01 | 36.37 | 29.48 | 35.31 | 28.76 |
| | HeadKV-R2 | 24.42 | 44.48 | **32.45** | 19.58 | 20.60 | 11.57 | **34.86** | 25.93 | 42.88 | 31.05 | **44.01** | 30.17 |
| | REAL | **26.40** | 45.49 | 32.23 | **23.45** | **24.47** | 17.79 | 33.66 | **29.50** | 42.88 | **33.35** | 40.51 | **31.79** |

Table 17: Question-aware, individual results of LongBench v2 (Bai et al., 2025) for Mistral-Large-Instruct-2411 (Mistral AI, 2024).

| Token Budget | Method | narrativeqa | qasper | multi-fieldqa | hotpotqa | 2wikimqa | musique | gov-report | qmsum | multi-news | trec | triviaqa | samsum | passagecnt | pass-ret. | lcc | repobench-p | Avg. |
|---|---|---|---|---|---|---|---|---|---|---|---|---|---|---|---|---|---|---|
| N/A | Full-KV | 25.56 | 32.07 | 39.71 | 43.57 | 35.28 | 21.18 | 28.71 | 23.26 | 26.64 | 73.50 | 90.48 | 42.33 | 4.80 | 69.25 | 59.29 | 54.05 | 41.86 |
| 0.1 | DuoAttention | 6.78 | 6.10 | 11.40 | 11.51 | 12.46 | 2.46 | 7.58 | 4.00 | 3.55 | 29.84 | 31.92 | 9.77 | 2.67 | 20.28 | 21.21 | 12.19 | 12.11 |
| | SnapKV | 11.45 | 11.64 | 19.26 | 14.20 | 14.54 | 9.93 | 9.79 | 11.56 | 9.99 | 32.05 | 28.41 | 19.39 | 1.87 | 29.29 | 21.17 | 25.46 | 16.87 |
| | PyramidKV | 9.99 | 15.20 | 13.42 | 19.69 | 13.02 | 8.58 | 13.87 | 7.37 | 11.88 | 27.34 | 45.06 | 14.60 | 2.24 | 27.28 | 25.55 | 17.78 | 17.05 |
| | HeadKV-R | 12.35 | 12.55 | 20.43 | 17.16 | 17.77 | 9.90 | 16.30 | 10.26 | 13.35 | **37.35** | 32.04 | 21.18 | 3.91 | 35.17 | 24.77 | 24.65 | 19.32 |
| | HeadKV-R2 | 14.16 | 18.90 | 17.01 | 25.54 | 17.27 | 13.54 | **18.60** | 12.27 | 16.70 | 33.09 | 36.71 | **24.48** | 6.42 | 27.21 | 22.07 | 25.96 | 20.62 |
| | REAL | **15.66** | **22.30** | **22.65** | **26.87** | **18.47** | **17.36** | 17.54 | **16.58** | **20.13** | 32.21 | **49.34** | 23.13 | **8.71** | **37.40** | **25.74** | **33.05** | **24.20** |
| 0.2 | DuoAttention | 8.08 | 20.76 | 15.97 | 27.64 | 17.57 | 8.95 | 17.77 | 14.32 | 9.66 | 18.80 | 19.19 | 16.23 | 3.16 | 23.60 | 17.53 | 25.40 | 16.54 |
| | SnapKV | 9.28 | 18.86 | 18.38 | 24.57 | 23.22 | 12.73 | 20.32 | 16.61 | 13.08 | 24.48 | 15.34 | 22.14 | 3.64 | 27.36 | 21.06 | 31.43 | 20.16 |
| | PyramidKV | 11.97 | 24.86 | 23.32 | 32.56 | 23.54 | 11.94 | 20.84 | 18.41 | 13.23 | 32.91 | 29.87 | 22.96 | 3.75 | 35.03 | 26.24 | 35.18 | 22.91 |
| | HeadKV-R | 12.76 | 24.41 | 23.11 | 31.28 | 26.82 | 16.69 | 23.91 | 20.35 | **17.02** | 33.01 | 32.89 | **27.72** | 4.50 | 38.23 | 27.52 | 40.51 | 25.05 |
| | HeadKV-R2 | 15.51 | 32.14 | 27.34 | 34.81 | 26.66 | 16.58 | 23.41 | 19.93 | 13.81 | 42.93 | 36.64 | 25.96 | 3.57 | 42.50 | 37.86 | 42.14 | 27.61 |
| | REAL | **22.03** | **33.83** | **29.92** | **37.42** | **29.84** | **17.72** | **26.00** | **20.89** | 16.46 | **42.95** | **42.37** | 27.63 | **4.64** | **45.27** | **39.28** | **44.19** | **30.03** |
| 0.4 | DuoAttention | 14.66 | 21.30 | 21.65 | 25.87 | 17.47 | 16.36 | 16.54 | 15.58 | 19.13 | 31.21 | 48.34 | 22.13 | 2.71 | 36.40 | 24.74 | 32.05 | 22.88 |
| | SnapKV | 18.76 | 24.37 | 29.10 | 33.25 | 25.95 | 15.90 | 21.37 | 16.49 | 19.86 | 55.37 | 70.93 | 31.85 | 2.64 | 53.57 | 44.12 | 41.86 | 31.59 |
| | PyramidKV | 19.76 | 25.37 | 30.10 | 34.25 | 26.95 | 16.90 | 22.37 | 17.49 | 20.86 | 56.37 | 71.93 | 32.85 | 3.64 | 54.57 | 45.12 | 42.86 | 32.59 |
| | HeadKV-R | 21.04 | 24.53 | 35.30 | 31.11 | 30.23 | 16.77 | 21.19 | 21.31 | 20.59 | 61.81 | 65.69 | 37.84 | 3.91 | 52.56 | 51.70 | 43.56 | 33.70 |
| | HeadKV-R2 | 20.69 | 28.67 | 31.90 | 37.29 | 32.28 | 20.53 | 27.02 | 21.31 | 26.52 | 53.16 | 71.5 | 34.52 | 5.00 | 60.00 | 55.72 | 46.23 | 35.77 |
| | REAL | **23.76** | **30.08** | **37.04** | **43.10** | **41.32** | **21.49** | **27.74** | **22.76** | **26.54** | **73.47** | **89.81** | **46.87** | **5.00** | **66.00** | **57.30** | **51.59** | **41.49** |

Table 18: Question-agnostic, individual results of LongBench (Bai et al., 2024) for Llama-3-8B-Instruct (Grattafiori et al., 2024).

| Token Budget | Method | narrativeqa | qasper | multi-fieldqa | hotpotqa | 2wikimqa | musique | gov-report | qmsum | multi-news | trec | triviaqa | samsum | passagecnt | pass-ret. | lcc | repobench-p | Avg. |
|---|---|---|---|---|---|---|---|---|---|---|---|---|---|---|---|---|---|---|
| N/A | Full-KV | 26.63 | 32.99 | 49.34 | 42.77 | 27.35 | 18.78 | 32.87 | 23.24 | 27.10 | 71.00 | 86.23 | 42.79 | 2.75 | 86.98 | 56.93 | 54.49 | 42.64 |
| 0.1 | DuoAttention | 8.66 | 15.30 | 15.65 | 19.87 | 11.47 | 10.36 | 10.54 | 9.58 | 13.13 | 25.21 | 42.34 | 16.13 | 1.71 | 30.40 | 18.74 | 26.05 | 17.20 |
| | SnapKV | 14.81 | 16.35 | 21.95 | 18.68 | 18.75 | 14.08 | 14.31 | 14.26 | 14.44 | 27.81 | 48.06 | 19.58 | 6.09 | 32.12 | 24.28 | 29.02 | 20.91 |
| | PyramidKV | 13.11 | 15.62 | 21.13 | 20.43 | 18.49 | 10.40 | 15.73 | 11.05 | 13.48 | 38.88 | 43.70 | 21.88 | 2.39 | 37.05 | 27.98 | 27.51 | 21.17 |
| | HeadKV-R | 14.98 | 19.22 | 21.33 | 25.92 | 19.35 | 13.34 | 15.70 | 14.84 | 15.75 | 38.88 | 49.49 | 23.05 | 4.97 | 34.97 | 32.82 | 28.94 | 23.35 |
| | HeadKV-R2 | 15.02 | 20.13 | 21.89 | 33.70 | 20.63 | 20.81 | 27.63 | 21.44 | 14.30 | 52.39 | 52.28 | 26.74 | 4.50 | 38.50 | 37.69 | 39.12 | 27.92 |
| | REAL | **17.32** | **22.98** | **24.19** | **36.57** | **30.02** | **22.74** | **29.54** | **21.89** | **17.03** | **62.56** | **69.59** | **32.17** | **6.00** | **53.50** | **44.65** | **43.67** | **33.40** |
| 0.2 | DuoAttention | 11.78 | 11.10 | 16.40 | 16.51 | 17.46 | 7.46 | 12.58 | 9.00 | 8.55 | 34.84 | 36.92 | 14.77 | 2.33 | 25.28 | 26.21 | 17.19 | 16.77 |
| | SnapKV | 14.09 | 21.36 | 22.95 | 33.26 | 20.35 | 11.66 | 19.41 | 10.70 | 16.96 | 52.43 | 47.62 | 26.21 | 5.27 | 44.69 | 42.56 | 41.40 | 26.93 |
| | PyramidKV | 13.04 | 16.53 | 27.30 | 23.11 | 22.23 | 8.77 | 13.19 | 13.31 | 12.59 | 53.81 | 57.69 | 29.84 | 4.09 | 44.56 | 43.70 | 35.56 | 25.57 |
| | HeadKV-R | 19.35 | 19.55 | 27.43 | 24.16 | 24.77 | 16.90 | 23.30 | 17.26 | 20.35 | 44.35 | 39.04 | 28.18 | 4.23 | 42.17 | 31.77 | 31.65 | 25.92 |
| | HeadKV-R2 | 24.36 | 27.89 | 36.03 | 35.76 | 26.32 | **23.79** | 27.00 | **24.99** | 16.01 | 62.73 | 75.25 | 36.58 | 5.00 | 57.27 | 47.31 | 45.16 | 35.03 |
| | REAL | **26.38** | **31.75** | **47.68** | **41.47** | **26.68** | 18.27 | **29.37** | 21.52 | **28.61** | **73.29** | **85.44** | **41.71** | **6.50** | **65.89** | **53.20** | **51.07** | **44.60** |

Table 19: Question-agnostic, individual results of LongBench (Bai et al., 2024) for Mistral-7B-Ins-v0.2 (Jiang et al., 2023).

| Token Budget | Method | narrativeqa | qasper | multi-fieldqa | hotpotqa | 2wikimqa | musique | gov-report | qmsum | multi-news | trec | triviaqa | samsum | passagecnt | pass-ret. | lcc | repobench-p | Avg. |
|---|---|---|---|---|---|---|---|---|---|---|---|---|---|---|---|---|---|---|
| 64 | +RR, –D, –B | 18.72 | 9.99 | 19.82 | 25.18 | 15.33 | 10.29 | 16.94 | 9.16 | 12.90 | 44.75 | 52.95 | 22.92 | 4.91 | 29.25 | 35.74 | 40.48 | 22.37 |
| | +RR, +D, –B | 18.80 | 15.72 | 27.90 | 33.77 | 23.01 | 15.28 | 17.02 | 14.40 | 18.16 | 60.02 | 79.45 | 34.03 | 4.94 | 46.01 | 50.32 | 54.30 | 33.58 |
| | +RR, –D, +B | 18.69 | 16.06 | 23.96 | 32.99 | 23.44 | 15.89 | 16.92 | 14.72 | 15.60 | 58.64 | 80.97 | 35.38 | 4.91 | 47.02 | 43.23 | 53.05 | 34.22 |
| | +RR, +D, +B | **22.02** | **23.72** | **32.88** | **37.11** | **26.21** | **17.52** | **19.93** | **21.73** | **21.40** | **65.96** | **90.53** | **39.01** | **5.78** | **69.42** | **59.31** | **59.67** | **38.26** |
| 128 | +RR, –D, –B | 25.15 | 23.56 | 34.87 | 38.86 | 25.33 | 17.55 | 23.15 | 17.95 | 22.66 | 65.54 | 66.50 | 34.22 | 5.19 | 53.70 | 55.04 | 52.79 | 30.69 |
| | +RR, +D, –B | 24.11 | 25.82 | 37.02 | 40.35 | 29.13 | 17.85 | 22.19 | 19.67 | 24.06 | 68.05 | 76.50 | 34.80 | 4.98 | 58.84 | 58.44 | 54.80 | 35.30 |
| | +RR, –D, +B | 25.13 | 28.27 | 35.87 | 39.27 | 31.42 | 20.25 | 23.13 | 21.54 | 23.31 | 66.23 | 82.51 | 39.49 | **5.19** | 64.43 | 56.62 | 53.36 | 38.07 |
| | +RR, +D, +B | **24.69** | **29.49** | **37.04** | **43.36** | **33.37** | **20.07** | **22.73** | **22.47** | **24.07** | **73.12** | **87.62** | **39.13** | 5.10 | **67.21** | **58.47** | **58.89** | **40.43** |

Table 20: Question-aware, behavior ablation on LongBench (Bai et al., 2024) with Llama-3-8B-Instruct (Grattafiori et al., 2024).

| Token Budget | Method | narrativeqa | qasper | multi-fieldqa | hotpotqa | 2wikimqa | musique | gov-report | qmsum | multi-news | trec | triviaqa | samsum | passagecnt | pass-ret. | lcc | repobench-p | Avg. |
|---|---|---|---|---|---|---|---|---|---|---|---|---|---|---|---|---|---|---|
| N/A | Full-KV | 25.56 | 32.07 | 39.71 | 43.57 | 35.28 | 21.18 | 28.71 | 23.26 | 26.64 | 73.50 | 90.48 | 42.33 | 4.80 | 69.25 | 59.29 | 54.05 | 41.86 |
| 64 | Random_score | 12.76 | 9.47 | 21.69 | 18.86 | 13.00 | 8.61 | 11.55 | 8.68 | 14.12 | 33.52 | 44.89 | 19.17 | 3.35 | 27.72 | 39.13 | 30.32 | 18.97 |
| | Mean_score | 16.63 | 12.81 | 26.36 | 24.02 | 16.96 | 11.40 | 15.06 | 11.73 | 17.15 | 42.69 | 58.59 | 25.38 | 4.37 | 37.48 | 47.54 | 38.61 | 24.76 |
| | Max_score | 18.06 | 16.60 | 28.69 | 27.18 | 20.68 | 13.34 | 16.34 | 15.20 | 18.67 | 48.32 | 71.41 | 29.71 | 4.74 | 48.57 | 51.75 | 43.70 | 30.18 |
| | REAL | **22.02** | **23.72** | **32.88** | **37.11** | **26.21** | **17.52** | **19.93** | **21.73** | **21.40** | **65.96** | **90.53** | **39.01** | **5.78** | **69.42** | **59.31** | **59.67** | **38.26** |
| 128 | Random_score | 11.11 | 15.39 | 21.24 | 20.62 | 14.92 | 10.19 | 5.21 | 5.01 | 4.07 | 15.30 | 32.09 | 17.60 | 2.66 | 38.54 | 27.80 | 26.33 | 20.53 |
| | Mean_score | 17.14 | 20.98 | 26.63 | 31.91 | 27.68 | 15.01 | 15.21 | 14.61 | 18.14 | 51.56 | 64.47 | 27.16 | 3.63 | 48.33 | 43.02 | 48.83 | 30.24 |
| | Max_score | 19.57 | 24.42 | 29.17 | 34.67 | 30.40 | 17.15 | 18.26 | 19.56 | 20.77 | 59.14 | 74.12 | 31.02 | 4.22 | 52.94 | 46.76 | 53.64 | 34.56 |
| | REAL | **24.69** | **29.49** | **37.04** | **43.36** | **33.37** | **20.07** | **22.73** | **22.47** | **24.07** | **73.12** | **87.62** | **39.13** | **5.10** | **67.21** | **58.47** | **58.89** | **40.43** |

Table 21: Question-aware, metric ablation on LongBench (Bai et al., 2024) with Llama-3-8B-Instruct (Grattafiori et al., 2024).